\documentclass[10pt,twocolumn,letterpaper]{article}

\usepackage[pagenumbers]{cvpr} 

\usepackage{graphicx}
\usepackage{amsmath}
\usepackage{amssymb}
\usepackage{booktabs}

\makeatletter
\@namedef{ver@everyshi.sty}{}
\makeatother

\usepackage{array}
\usepackage{subcaption}
\usepackage{multirow,multicol}
\usepackage[flushleft]{threeparttable}
\usepackage{comment}
\usepackage{algorithmic}
\usepackage{booktabs}
\usepackage{bbm, dsfont}
\usepackage[font=small,labelfont=bf]{caption}

\usepackage{amssymb}
\usepackage{pifont}

\newcommand{\cmark}{\text{\ding{51}}}%
\newcommand{\xmark}{\text{\ding{55}}}%

\newcommand{\tablestyle}[2]{\setlength{\tabcolsep}{#1}\renewcommand{\arraystretch}{#2}\centering\footnotesize}

\newcolumntype{x}[1]{>{\centering\arraybackslash}p{#1pt}}
\newcommand{\app}{\raise.17ex\hbox{$\scriptstyle\sim$}}

\newlength\savewidth\newcommand\shline{\noalign{\global\savewidth\arrayrulewidth
  \global\arrayrulewidth 1pt}\hline\noalign{\global\arrayrulewidth\savewidth}}

\makeatletter\renewcommand\paragraph{\@startsection{paragraph}{4}{\z@}
  {.5em \@plus1ex \@minus.2ex}{-.5em}{\normalfont\normalsize\bfseries}}\makeatother

\def\tablecite#1#{%
  \def\pretablecite{#1}%
  \tableciteaux}
\def\tableciteaux#1{%
  \textsuperscript{\expandafter\originalcite\pretablecite{#1}}%
}

\usepackage{graphicx}
\usepackage{enumitem}
\usepackage{wrapfig}
\usepackage{lipsum}
\usepackage[table]{xcolor}
\usepackage{soul}

\newcolumntype{H}{>{\setbox0=\hbox\bgroup}c<{\egroup}@{}}
\newcolumntype{a}{>{\columncolor{Gray}}c}
\DeclareRobustCommand{\colorrowtext}[0]{{\sethlcolor{Gray}\hl{gray}}}
\usepackage{tabu}
\usepackage{xcolor}
\usepackage{nicematrix}

\definecolor{ForestGreen}{rgb}{0.13, 0.55, 0.13}
\definecolor{Green}{rgb}{0.0, 0.5, 0.0}
\definecolor{green(munsell)}{rgb}{0.0, 0.66, 0.47}
\definecolor{green(ryb)}{rgb}{0.4, 0.69, 0.2}
\definecolor{green(pigment)}{rgb}{0.0, 0.65, 0.31}
\definecolor{citecolor}{HTML}{0071bc}
\definecolor{GrayXMark}{gray}{0.7}
\newcommand{\grayxmark}{ {\color{GrayXMark} \ding{55}} } %
\newcommand{\graydash}{ {\color{GrayXMark} -} } %
\usepackage[pagebackref,breaklinks,colorlinks,citecolor=citecolor]{hyperref}

\usepackage{tabularx}
\usepackage[export]{adjustbox}

\usepackage[capitalize]{cleveref}
\crefname{section}{Sec.}{Secs.}
\Crefname{section}{Section}{Sections}
\Crefname{table}{Table}{Tables}
\crefname{table}{Table}{Tabs.}

\newcommand{\Ours}{CutLER\xspace}

\newcommand{\imnet}{ImageNet\xspace}
\newcommand{\maskcut}{MaskCut\xspace}
\newcommand{\droploss}{DropLoss\xspace}
\newcommand{\pretraining}{pretraining\xspace}

\definecolor{ForestGreen}{rgb}{0.13, 0.55, 0.13}
\definecolor{Green}{rgb}{0.0, 0.5, 0.0}
\definecolor{green(munsell)}{rgb}{0.0, 0.66, 0.47}
\definecolor{green(ryb)}{rgb}{0.4, 0.69, 0.2}
\definecolor{green(pigment)}{rgb}{0.0, 0.65, 0.31}

\usepackage{tikz}

\def\cvprPaperID{1167} 
\def\confName{CVPR}
\def\confYear{2023}

\title{Cut and Learn for Unsupervised Object Detection and Instance Segmentation}

\author{
  Xudong Wang$^{1,2}$ \quad \quad
  Rohit Girdhar$^{1}$ \quad \quad
  Stella X. Yu$^{2,3}$ \quad \quad
  Ishan Misra$^{1}$ \\
  $^{1}$FAIR, Meta AI \quad \quad $^{2}$UC Berkeley / ICSI \quad \quad $^{3}$University of Michigan \\
  \small{Code:} \href{https://github.com/facebookresearch/CutLER}{\small{https://github.com/facebookresearch/CutLER}}
  }

\begin{document}

\twocolumn[{%
  \renewcommand\twocolumn[1][]{#1}%
  \maketitle
    \vspace{-20pt}
    \captionsetup{type=figure}
    \centering
    \includegraphics[width=0.98\textwidth]{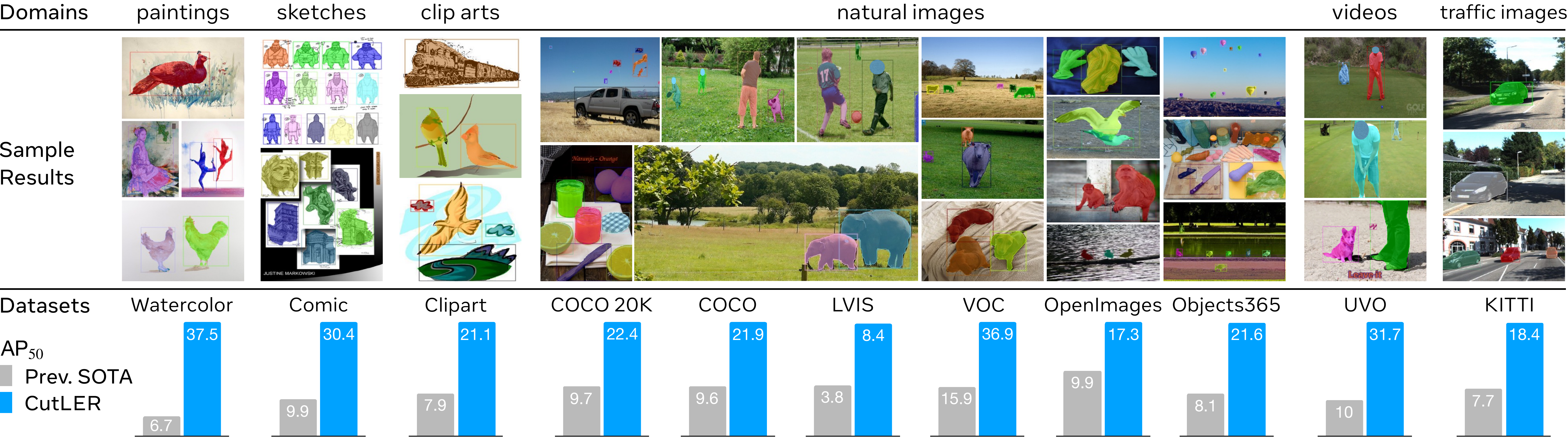}
    \vspace{-4pt}
    \caption{\textbf{Zero-shot unsupervised object detection and instance segmentation} using our \Ours model, which is trained without human supervision.
    We evaluate the model using the standard detection AP$^{\text{box}}_{50}$.
    \Ours gives a strong performance on a variety of benchmarks spanning diverse image domains - video frames, paintings, clip arts, complex scenes, \etc.
    Compared to the previous state-of-the-art method, FreeSOLO~\cite{wang2022freesolo} with a backbone of ResNet101, \Ours with a backbone of ResNet50 provides strong gains on all benchmarks, increasing performance by more than $2\times$ on 10 of the 11 benchmarks.
    We evaluate~\cite{wang2022freesolo} with its official code and checkpoint.
    }
    \label{fig:teaser}
    \vspace{14pt}
}]

\begin{abstract}
    We propose \textbf{Cut}-and-\textbf{LE}a\textbf{R}n (\Ours), a simple approach for training unsupervised object detection and segmentation models.
    We leverage the property of self-supervised models to `discover' objects without supervision and amplify it to train a state-of-the-art localization model without any human labels.
    \Ours first uses our proposed \maskcut approach to generate coarse masks for multiple objects in an image, and then learns a detector on these masks using our robust loss function.
    We further improve performance by self-training the model on its predictions.
    Compared to prior work, \Ours is simpler, compatible with different detection architectures, and detects multiple objects.
    \Ours is also a zero-shot unsupervised detector and improves detection performance AP$_{50}$ by over 2.7$\times$ on 11 benchmarks across domains like video frames, paintings, sketches, \etc
    With finetuning, \Ours serves as a low-shot detector surpassing MoCo-v2 by 7.3\% AP$^{\text{box}}$ and 6.6\% AP$^{\text{mask}}$ on COCO when training with 5\% labels.
\end{abstract}
\section{Introduction}
Object localization is a critical task in computer vision that enables AI systems to perceive, reason, plan and act in an object-centric manner.
Training models for localization require special annotations like object boxes, masks, localized points, \etc which are both difficult and resource intensive to collect.
Without accounting for overhead, annotating $\sim$164K images in the COCO dataset~\cite{lin2014microsoft} with masks for just 80 classes took more than 28K human hours of annotation time.
In this work, we study unsupervised object detection and instance segmentation models that can be trained without any human labels.
Our key insight is that simple probing and training mechanisms can amplify the innate localization ability of self-supervised models~\cite{caron2021emerging}, leading to state-of-the-art unsupervised zero-shot detectors.

Our method \textbf{Cut}-and-\textbf{LE}a\textbf{R}n (\Ours) consists of three simple, architecture- and data-agnostic mechanisms.
Consistent with prior self-supervised learning methods~\cite{he2020momentum,chen2020improved,chen2020simple,caron2021emerging}, \Ours is trained exclusively on unlabeled \imnet data without needing additional training data, but contrary to these methods, \Ours can be directly employed to perform complex segmentation and detection tasks over a wide range of domains.
\textit{First}, we propose \maskcut that can automatically produce \emph{multiple} initial coarse masks for each image, using the pretrained self-supervised features.
\textit{Second}, we propose a simple loss dropping strategy to train detectors using the coarse masks while being robust to objects missed by \maskcut.
\textit{Finally}, we observe that despite learning from these coarse masks, the detectors `clean' the ground truth and produce masks (and boxes) that are better than the coarse masks used to train them.
Therefore, we further show that multiple rounds of self-training on the models' own predictions allow it to evolve from capturing the similarity of local pixels to capturing the global geometry of the object, thus producing finer segmentation masks.

Prior work shows that a self-supervised vision transformer (ViT)~\cite{dosovitskiy2020image} can automatically learn patch-wise features that detect a single \emph{salient} object in an image~\cite{caron2021emerging,vo2020toward,vo2021large,simeoni2021localizing,wang2022tokencut}.
However, unlike \Ours, such salient object detection methods only locate a single, usually the most prominent, object and cannot be used for real world images containing multiple objects.
While some recent methods, \eg, FreeSOLO~\cite{wang2022freesolo} and DETReg~\cite{bar2022detreg}, also aim at unsupervised multi-object detection (or multi-object discovery), they rely on a particular detection architecture, \eg, SOLO-v2~\cite{wang2020solov2} or DDETR~\cite{carion2020end,zhu2020deformable}.
Additionally, apart from self-supervised features trained on \imnet~\cite{deng2009imagenet}, the current state-of-the-art methods FreeSOLO and MaskDistill~\cite{van2022discovering} also require `in-domain' unlabeled data for model training.

In contrast, \Ours works with various detection architectures and can be trained solely on \imnet, without requiring in-domain unlabeled data.
Thus, during model training, \Ours does not see any images from any target dataset and yields a zero-shot model capable of detecting and segmenting multiple objects in diverse domains.

\noindent \textbf{Features of \Ours.}
\textit{\textbf{1) Simplicity:}} \Ours is simple to train and agnostic to the choice of detection and backbone architectures. Thus, it can be integrated effortlessly into existing object detection and instance segmentation works.
\textit{\textbf{2) Zero-shot detector:}} \Ours trained solely on ImageNet shows strong zero-shot performance on 11 different benchmarks where it outperforms prior work trained with additional in-domain data.
We \textbf{double the AP$^{\text{box}}_{50}$} performance on 10 of these benchmarks, as shown in~\cref{fig:teaser}, and even outperform supervised detectors on the UVO video instance segmentation benchmark.
\textit{\textbf{3) Robustness:}} \Ours exhibits strong robustness against domain shifts when tested on images from different domains such as video frames, sketches, paintings, clip arts, \etc.
\textit{\textbf{4) Pretraining for supervised detection:}} \Ours can also serve as a pretrained model for training fully supervised object detection and instance segmentation models and improves performance on COCO, including on few-shot object detection benchmarks.
\def\tabDistMethods#1{
    \begin{table}[#1]
    \centering
    \tablestyle{1.0pt}{1.0}
    \small
    \begin{tabular}{l|ccccc}
    & DINO & LOST & TokenCut & FreeSOLO & Ours \\ \shline
    detect multiple objects & \grayxmark & \cmark & \grayxmark & \cmark & \cmark \\\hline
    zero-shot detector & \cmark & \grayxmark & \cmark & \grayxmark & \cmark \\\hline
    \begin{tabular}[c]{@{}l@{}}compatible with various \\ detection architectures \end{tabular} & \graydash & \cmark & \graydash & \grayxmark & \cmark \\\hline
    \begin{tabular}[c]{@{}l@{}}pretrained model for \\ supervised detection \end{tabular} & \cmark & \grayxmark & \grayxmark & \cmark & \cmark \\
    \hline
    \end{tabular}\vspace{-6pt}
    \caption{
    We compare previous methods on unsupervised object detection, including DINO~\cite{caron2021emerging}, LOST~\cite{simeoni2021localizing}, TokenCut~\cite{wang2022tokencut} and FreeSOLO~\cite{wang2022freesolo}, with our \Ours in term of key properties. 
    Our \Ours is the only method with all these desired properties.
    }
    \label{tab:distMethods}
    \end{table}
}
\newcolumntype{A}{>{\centering}p{0.03\textwidth}}
\newcolumntype{B}{p{0.04\textwidth}}
\newcommand{\green}[1]{\multicolumn{1}{c}{\color{green(pigment)}#1}}
\newcommand{\red}[1]{\multicolumn{1}{c}{\color{red}#1}}
\newcommand{\cell}[1]{\multicolumn{1}{r}{#1}}

\section{Related Work}
\textbf{Self-supervised feature learning} involves inferring the patterns within the large-scale unlabeled data without using human-annotated labels. {\textit{Contrastive learning based}} \cite{wu2018unsupervised, misra2020self,he2020momentum, chen2020simple} methods learn such representations that similar samples or various augmentations of the same instance are close to each other, while dissimilar instances are far apart.
{\textit{Similarity-based self-supervised learning}} methods \cite{grill2020bootstrap, chen2021exploring} learn representations via minimizing the distance between different augmentations of the same instance and use only positive sample pairs. 
{\textit{Clustering-based feature learning}} \cite{xie2016unsupervised, asano2019self,zhuang2019local, caron2020unsupervised, wang2021unsupervised} automatically discovers the natural grouping of data in the latent representation space.
Recently, \cite{he2022masked, bao2021beit} have shown that {\textit{masked autoencoders}}, which learn representations via masking out a large random subset of image patches and reconstructing the missing pixels or patches~\cite{doersch2015unsupervised, doersch2014context, he2022masked, bao2021beit}, are scalable self-supervised learners for computer vision~\cite{he2022masked}.

In contrast to these unsupervised representation learning efforts,
our work aims to automatically discover natural pixel groupings and locate instances within each image.

\tabDistMethods{t!}
\textbf{Unsupervised object detection and instance segmentation}.
The main comparisons to previous works are listed in~\cref{tab:distMethods} and are elaborated as follows:

DINO~\cite{caron2021emerging} observes that the underlying semantic segmentation of images can emerge from the self-supervised Vision Transformer (ViT)~\cite{dosovitskiy2020image}, which does not appear explicitly in either supervised ViT or ConvNets~\cite{ziegler2022self,caron2021emerging}.
Based on this observation, LOST~\cite{simeoni2021localizing} and TokenCut~\cite{wang2022tokencut} leverage self-supervised ViT features and propose to segment \textit{one single} salient object~\cite{cho2015unsupervised,simeoni2021localizing,wang2022tokencut} from each image based on a graph that is constructed with DINO's patch features.

These previous works either can not detect more than one object from each image, \eg, DINO and TokenCut, or can not improve the quality of features for better transfer to downstream detection and segmentation tasks, \eg, TokenCut and LOST. Unlike these works, \Ours can locate multiple objects and serve as a pretrained model for label-efficient and fully-supervised learning.

FreeSOLO~\cite{wang2022freesolo} achieves unsupervised instance segmentation by extracting coarse object masks in an unsupervised manner, followed by mask refinement through a self-training procedure. 
While FreeSOLO's FreeMask stage can generate multiple coarse masks per image, the quality of these masks is often rather low~\cite{wang2022freesolo}.
MaskDistill~\cite{van2022discovering} distills class-agnostic initial masks from the affinity graph produced by a self-supervised DINO~\cite{caron2021emerging}. 
However, it utilizes \textit{one single} mask per image in the distillation stage, which greatly limits the model's ability to detect multiple objects.

By contrast, the initial masks generated by our \maskcut are usually better in quality and quantity than the initial masks used by~\cite{wang2022freesolo,van2022discovering}. Therefore, \Ours achieves 2$\times\app$4$\times$ higher AP$^\text{box}$ and AP$^\text{mask}$ than FreeSOLO~\cite{wang2022freesolo} and MaskDistill~\cite{van2022discovering} on almost all experimented detection and segmentation benchmarks, even when FreeSOLO and MaskDistill are trained and tested on the same domain.
\def\figMaskCut#1{
    \captionsetup[sub]{font=small}
    \begin{figure*}[#1]
      \centering
      \includegraphics[width=0.99\linewidth]{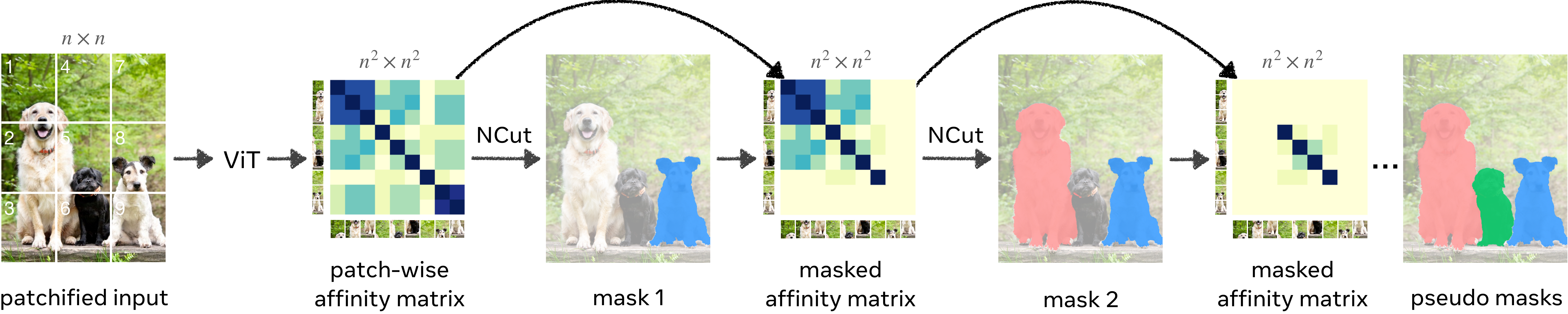}\vspace{-3pt}
      \caption{\textbf{\maskcut} can discover multiple object masks in an image without supervision.
      We build upon~\cite{caron2021emerging,wang2022tokencut} and create a patch-wise similarity matrix for the image using a self-supervised DINO~\cite{caron2021emerging} model's features.
      We apply Normalized Cuts~\cite{shi2000normalized} to this matrix and obtain a single foreground object mask of the image.
      We then mask out the affinity matrix values using the foreground mask and repeat the process, which allows \maskcut to discover multiple object masks in a single image. In this pipeline illustration, we set $n\!=\!3$.
      }
      \label{fig:MaskCut}
    \end{figure*}
}

\def\figPipeline#1{
    \captionsetup[sub]{font=small}
    \begin{figure}[#1]
      \centering
      \includegraphics[width=1.0\linewidth]{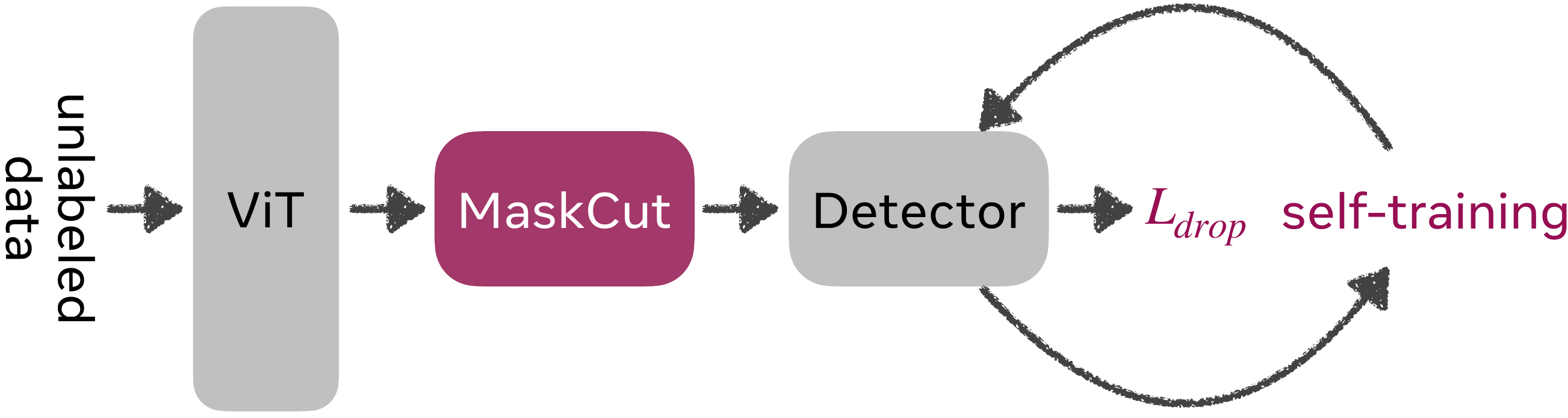}
      \caption{\textbf{Overview of \Ours.}
      We propose a simple yet effective method to train an object detection and instance segmentation model without using any supervision.
      We first propose \maskcut to extract initial coarse masks from the features of a self-supervised ViT.
      We then learn a detector using our loss dropping strategy that is robust to objects missed by \maskcut.
      We further improve the model using multiple rounds of self-training.
      }
      \label{fig:pipeline}
    \end{figure}
}

\newcommand{\plus}[1]{\small\bf\textcolor{Green}{#1}}
\newcommand{\spplus}[1]{\scriptsize\bf\textcolor{Green}{#1}}
\newcommand{\STAB}[1]{\begin{tabular}{@{}c@{}}#1\end{tabular}}
\newcommand{\ROTATED}[1]{\footnotesize \multirow{1}{*}{\STAB{\rotatebox[origin=c]{0}{#1}}}}

\def\tabZeroShotAllDatasets#1{
\begin{table*}[#1]
\tablestyle{0.45pt}{1.0}
\footnotesize
\begin{center}
\begin{tabular}{p{1.9cm}|cc|cc|cc|cc|cc|cc|cc|cc|cc|cc|cc|cc}
\multirow{1}{*}{Datasets $\rightarrow$} & \multicolumn{2}{c|}{\ROTATED{Avg.}} & \multicolumn{2}{c|}{\ROTATED{COCO}} & \multicolumn{2}{c|}{\ROTATED{COCO20K}} & \multicolumn{2}{c|}{\ROTATED{VOC}} & \multicolumn{2}{c|}{\ROTATED{LVIS}} & \multicolumn{2}{c|}{\ROTATED{UVO}} & \multicolumn{2}{c|}{\ROTATED{Clipart}} & \multicolumn{2}{c|}{\ROTATED{Comic}}  & \multicolumn{2}{c|}{\ROTATED{Watercolor}} & \multicolumn{2}{c|}{\ROTATED{KITTI}} & \multicolumn{2}{c|}{\ROTATED{Objects365}} & \multicolumn{2}{c}{\ROTATED{OpenImages}} \\
Metrics $\rightarrow$ & AP$_{50}$ & AR & AP$_{50}$ & AR & AP$_{50}$ & AR & AP$_{50}$ & AR & AP$_{50}$ & AR & AP$_{50}$ & AR & AP$_{50}$ & AR & AP$_{50}$ & AR & AP$_{50}$ & AR & AP$_{50}$ & AR & AP$_{50}$ & AR & AP$_{50}$ & AR \\
\shline
Prev. SOTA~\cite{wang2022freesolo} & \phantom{1}9.0 & 13.4 & \phantom{1}9.6 & 12.6 & \phantom{1}9.7 & 12.6 & 15.9 & 21.3 & 3.8 & \phantom{1}6.4 & 10.0 & 14.2 & \phantom{1}7.9 & 15.1 & \phantom{1}9.9 & 16.3 & \phantom{1}6.7 & 16.2 & \phantom{1}7.7 & \phantom{1}7.1 & \phantom{1}8.1 & 10.2 & \phantom{1}9.9 & 14.9 \\
\Ours & 24.3 & 35.5 & 21.9	& 32.7 & 22.4 & 33.1 & 36.9 & 44.3 & 8.4 & 21.8 & 31.7 & 42.8 & 21.1 & 41.3 & 30.4 & 38.6 & 37.5 & 44.6 & 18.4 & 27.5 & 21.6 & 34.2 & 17.3 & 29.6 \\
\textit{vs. prev. SOTA} & \spplus{+15.3} & \spplus{+22.1} & \spplus{+12.3}	& \spplus{+20.1} & \spplus{+12.7} & \spplus{+20.5} & \spplus{+21.0} & \spplus{+23.0} &	\spplus{+4.6}	& \spplus{+15.4} &	\spplus{+21.7} & \spplus{+28.6} &	\spplus{+13.2} & \spplus{+26.2} & \spplus{+20.5} & \spplus{+22.3} & \spplus{+30.8} & \spplus{+28.4} & \spplus{+10.7} & \spplus{+20.4} & \spplus{+13.5} & \spplus{+24.0} & \spplus{+7.4} & \spplus{+14.7} \\
\hline
\end{tabular}
\end{center}
\vspace{-14pt}
\caption{State-of-the-art \textbf{zero-shot unsupervised object detection} performance on 11 different datasets spanning a variety of domains. We report class-agnostic multi-object detection performance and the averaged results for 11 datasets using AP$^{\text{box}}_{50}$ and AR$^{\text{box}}_{100}$. Our \Ours is trained in an unsupervised manner solely on ImageNet. While the previous SOTA method~\cite{wang2022freesolo} is typically fine-tuned on extra data, \eg, $\sim$241k unlabeled COCO images, \Ours significantly outperforms it. Results of \cite{wang2022freesolo} are produced with official code and checkpoint.
}
\label{tab:zero-shot-all-datasets}
\end{table*}
}

\section{Method}
\label{sec:method}

\figPipeline{t!}
\figMaskCut{t!}

We tackle the problem of unsupervised object detection and segmentation with a simple cut-and-learn pipeline.
Our method builds upon insights from recent work~\cite{caron2021emerging,wang2022tokencut}, showing that self-supervised representations can discover objects.
While these methods often find a single object per image, we propose a simple approach that can discover multiple objects and significantly improves segmentation and detection performance.
The overview of our cut-and-learn pipeline is illustrated in~\cref{fig:pipeline}.
\textit{First}, we propose \maskcut that generates multiple binary masks per image using self-supervised features from DINO~\cite{caron2021emerging} (\cref{sec:masked-ncuts}).
\textit{Second}, we show a dynamic loss dropping strategy, called \droploss, that can learn a detector from \maskcut's initial masks while encouraging the model to explore objects missed by \maskcut (\cref{sec:weighted-loss});
\textit{Third}, we further improve the performance of our method through multiple rounds of self-training (\cref{sec:self-training}).

\subsection{Preliminaries}
\label{sec:preliminaries}
\noindent\textbf{Normalized Cuts} (NCut) treats the image segmentation problem as a graph partitioning task \cite{shi2000normalized}.
We construct a fully connected undirected graph via representing each image as a node.
Each pair of nodes is connected by edges with weights $W_{ij}$ that measure the similarity of the connected nodes.
NCut minimizes the cost of partitioning the graph into two sub-graphs, \ie, a bipartition, by solving a generalized eigenvalue system
\begin{align}
  (D-W)x = \lambda Dx
  \label{eqn:ncut}
\end{align}
\noindent for finding the eigenvector $x$ that corresponds to the second smallest eigenvalue $\lambda$, where $D$ is a $N\!\times\!N$ diagonal matrix with $d(i)=\sum_jW_{ij}$ and $W$ is a $N\!\times\!N$ symmetrical matrix.

\noindent \textbf{DINO and TokenCut.}
DINO~\cite{caron2021emerging} finds that the self-supervised ViT can automatically learn a certain degree of perceptual grouping of image patches.
TokenCut~\cite{wang2022tokencut} leverages the DINO features for NCut and obtaining foreground/background segments in an image.
The authors use the similarity of the patches in the DINO feature space as the similarity weight $W_{ij}$ in NCut.
Specifically, following multiple recent methods~\cite{simeoni2021localizing,wang2022tokencut,van2022discovering}, we use the cosine similarity of `key' features from the last attention layer of DINO-pretrained model, \ie, $W_{ij}\!=\!\frac{K_i K_j}{\|K_i\|_2 \|K_j\|_2}$ where $K_i$ is the `key' feature of patch $i$, and solve ~\cref{eqn:ncut} for finding the second smallest eigenvector $x$.

A limitation of TokenCut is that it only computes a single binary mask for an image and thus only finds one object per image.
Although we can use the other $N\!-\!2$ smallest eigenvectors to locate more than one instance, this significantly degrades the performance for multi-object discovery, as demonstrated in~\cref{sec:ablation}.

\subsection{MaskCut for Discovering Multiple Objects}

\label{sec:masked-ncuts}

As we discussed in~\cref{sec:preliminaries}, vanilla NCut is limited to discovering a single object in an image.
We propose \maskcut that extends NCut to discover multiple objects per image by iteratively applying NCut to a \emph{masked} similarity matrix (illustrated in~\cref{fig:MaskCut}).
After getting the bipartition $x^t$ from NCut at stage $t$, we get two disjoint groups of patches and construct a binary mask $M^t$, where
\begin{align}
  M^t_{ij}=
  \begin{cases}
    1, & \text{if } M^t_{ij}\geq \text{mean}(x^t)\\
    0, & \text{otherwise.}
  \end{cases}
  \label{eqn:binary-mask}
\end{align}
To determine which group corresponds to the foreground, we make use of two criteria: \textit{1)} intuitively, the foreground patches should be more prominent than background patches~\cite{vo2020toward,wang2022tokencut,caron2021emerging}. Therefore, the foreground mask should contain the patch corresponding to the maximum \textit{absolute} value in the second smallest eigenvector $M^t$; \textit{2)} we incorporate a simple but empirically effective object-centric prior~\cite{maji2011biased}: the foreground set should contain less than two of the four corners.
We reverse the partitioning of the foreground and background, \ie, $M^t_{ij}\!=\!1\!-\!M^t_{ij}$, if the criteria 1 is not satisfied while the current foreground set contains two corners or the criteria 2 is not satisfied.
In practice, we also set all $W_{ij}\!<\!\tau^{\text{ncut}}$ to $1e^{-5}$ and $W_{ij}\!\geq\!\tau^{\text{ncut}}$ to 1.

To get a mask for the $(t+1)^{\mathrm{th}}$ object, we update the node similarity $W^{t+1}_{ij}$ via masking out these nodes corresponding to the foreground in previous stages:
\begin{align}
  W^{t+1}_{ij}\!=\!\frac{(K_i\prod_{s=1}^{t}\hat{M}^s_{ij})(K_j\prod_{s=1}^{t}\hat{M}^s_{ij})}{\|K_i\|_2\|K_j\|_2}
  \label{eqn:update-graph}
\end{align}
\noindent where $\hat{M}^s_{ij}\!=\!1\!-\!{M}^s_{ij}$. Using the updated $W^{t+1}_{ij}$, we repeat~\cref{eqn:ncut,eqn:binary-mask} to get a mask $M^{t+1}$.
We repeat this process $t$ times and set $t\!=\!3$ by default.

\subsection{\droploss for Exploring Image Regions}
\label{sec:weighted-loss}
A standard detection loss penalizes predicted regions $r_i$ that do not overlap with the `ground-truth'.
Since the `ground-truth' masks given by \maskcut may miss instances, the standard loss does not enable the detector to discover new instances not labeled in the `ground-truth'.
Therefore, we propose to ignore the loss of predicted regions $r_i$ that have a small overlap with the `ground-truth'.
More specifically, during training, we drop the loss for each predicted region $r_i$ that has a maximum overlap of $\tau^{\text{IoU}}$ with any of the `ground-truth' instances:
\begin{align}
  \mathcal{L}_{\text{drop}}(r_i) = \mathbbm{1}(\text{IoU}_i^{\text{max}} > \tau^{\text{IoU}})\mathcal{L}_{\text{vanilla}}(r_i)
\end{align}
\noindent where $\text{IoU}_i^{\text{max}}$ denotes the maximum IoU with all `ground-truth' for $r_i$ and $\mathcal{L}_{\text{vanilla}}$ refers to the vanilla loss function of detectors.
$\mathcal{L}_{\text{drop}}$ does not penalize the model for detecting objects missed in the `ground-truth' and thus encourages the exploration of different image regions.
In practice, we use a low threshold $\tau^{\text{IoU}}=0.01$.

\subsection{Multi-Round Self-Training}
\label{sec:self-training}
Empirically, we find that despite learning from the coarse masks obtained by \maskcut, detection models `clean' the ground truth and produce masks (and boxes) that are better than the initial coarse masks used for training.
The detectors refine mask quality, and our \droploss strategy encourages them to discover new object masks.
Thus, we leverage this property and use multiple rounds of self-training to improve the detector's performance.

We use the predicted masks and proposals with a confidence score over $0.75\!-\!0.5t$ from the $t^\mathrm{th}$-round as the additional pseudo annotations for the $(t+1)^\mathrm{th}$-round of self-training.
To de-duplicate the predictions and the ground truth from round $t$, we filter out ground-truth masks with an IoU $>0.5$ with the predicted masks.
We found that three rounds of self-training are sufficient to obtain good performance.
Each round steadily increases the number of `ground-truth' samples used to train the model.

\subsection{Implementation Details}

\par \noindent \textbf{Training data.} We only use the images from the \imnet~\cite{deng2009imagenet} dataset (1.3 million images) for all parts of the \Ours model and do not use any type of annotations either for training or any supervised pretrained models.
\par \noindent \textbf{\maskcut.} We use \maskcut with three stages on images resized to 480$\times$480 pixels and compute a patch-wise affinity matrix using the ViT-B/8~\cite{dosovitskiy2020image} DINO~\cite{caron2021emerging} model.
We use Conditional Random Field (CRF)~\cite{krahenbuhl2011efficient} to post-process the masks and compute their bounding boxes.
\par \noindent \textbf{Detector.} While \Ours is agnostic to the underlying detector, we use popular Mask R-CNN~\cite{he2017mask} and Cascade Mask R-CNN~\cite{cai2018cascade} for all experiments, and use Cascade Mask R-CNN by default, unless otherwise noted.
We train the detector on \imnet with initial masks and bounding boxes for $160$K iterations with a batch size of 16.
When training the detectors with a ResNet-50 backbone~\cite{he2016deep}, we initialize the model with the weights of a self-supervised pretrained DINO~\cite{caron2021emerging} model.
We explored other pre-trained models, including MoCo-v2~\cite{chen2020improved}, SwAV~\cite{caron2020unsupervised}, and CLD~\cite{wang2021unsupervised}, and found that they gave similar detection performance.
We also leverage the copy-paste augmentation~\cite{ghiasi2021simple,dwibedi2017cut} during the model training process. Rather than using the vanilla copy-paste augmentation, to improve the model's ability to segment small objects, we randomly downsample the mask with a scalar uniformly sampled between 0.3 and 1.0.
We then optimize the detector for $160$K iterations using SGD with a learning rate of 0.005, which is decreased by 5 after $80$K iterations, and a batch size of 16.
We apply a weight decay of $5\!\times\!10^{-5}$ and a momentum of 0.9.

\tabZeroShotAllDatasets{!t}

\par \noindent \textbf{Self-training.}
We initialize the detection model in each stage using the weights from the previous stage. We optimize the detector using SGD with a learning rate of 0.01 for 80$K$ iterations. Since the self-training stage can provide a sufficient number of pseudo-masks for model training, we don't use the DropLoss during the self-training stages.

We provide more details on model implementation and training in~\cref{appendix:training-details}.
\def\tabCOCOplusTwntyK#1{
\begin{table*}[#1]
\tablestyle{1.7pt}{1.0}
\small
\begin{center}
\begin{tabular}{p{2.1cm}ccccccccclcccccc}
\multirow{2}{*}{Methods} & \multirow{2}{*}{Pretrain} & \multirow{2}{*}{Detector} & \multirow{2}{*}{Init.}  & \multicolumn{6}{c}{{COCO 20K}} && \multicolumn{6}{c}{{COCO val2017}} \\
\cline{5-10} \cline{12-17}
& & & & \multicolumn{1}{c}{AP$_{50}^{\text{box}}$} & \multicolumn{1}{c}{AP$_{75}^{\text{box}}$} & \multicolumn{1}{c}{AP$^{\text{box}}$} & \multicolumn{1}{c}{AP$_{50}^{\text{mask}}$} & \multicolumn{1}{c}{AP$_{75}^{\text{mask}}$} & \multicolumn{1}{c}{AP$^{\text{mask}}$}
&& \multicolumn{1}{c}{AP$_{50}^{\text{box}}$} & \multicolumn{1}{c}{AP$_{75}^{\text{box}}$} & \multicolumn{1}{c}{AP$^{\text{box}}$} & \multicolumn{1}{c}{AP$_{50}^{\text{mask}}$} & \multicolumn{1}{c}{AP$_{75}^{\text{mask}}$} & \multicolumn{1}{c}{AP$^{\text{mask}}$} \\ [.1em]
\shline
\multicolumn{4}{l}{\bf \textit{non zero-shot methods}} &&&&&&& \\
LOST~\cite{simeoni2021localizing} & IN+COCO & FRCNN & DINO &  \phantom{1} - &  \phantom{1} - &  \phantom{1} - &  \phantom{1}2.4 &  \phantom{1}1.0 &  \phantom{1}1.1 &&  \phantom{1} - &  \phantom{1} - &  \phantom{1} - &  \phantom{1} - &  \phantom{1} - &  \phantom{1} -\\
MaskDistill~\cite{van2022discovering} & IN+COCO & MRCNN & MoCo &  \phantom{1} - &  \phantom{1} - &  \phantom{1} - &  \phantom{1}6.8 &  \phantom{1}2.1 &  \phantom{1}2.9 &&  \phantom{1} - &  \phantom{1} - &  \phantom{1} - &  \phantom{1} - &  \phantom{1} - &  \phantom{1} - \\
FreeSOLO$^*$~\cite{wang2022freesolo} & IN+COCO & SOLOv2 & DenseCL & \phantom{1}\phantom{1}9.7	&	 \phantom{1}3.2	&	 \phantom{1}4.1	&	 \phantom{1}{9.7} &  \phantom{1}{3.4} &	\phantom{1}{4.3} && \phantom{1}\phantom{1}{9.6}	&	\phantom{1} 3.1	&	\phantom{1}{4.2}	&	\phantom{1}9.4	&	\phantom{1}{3.3}	&	 \phantom{1}{4.3}\\
\hline
\multicolumn{4}{l}{\bf \textit{zero-shot methods}} &&&&&&& \\
DETReg~\cite{bar2022detreg} & IN & DDETR & SwAV & \phantom{1} - &  \phantom{1} - &  \phantom{1} - &  \phantom{1} - &  \phantom{1} - &  \phantom{1} - &&  \phantom{1}\phantom{1}3.1 &  \phantom{1}0.6 &  \phantom{1}1.0 &  \phantom{1}8.8 & \phantom{1}1.9 &  \phantom{1}3.3 \\
DINO~\cite{caron2021emerging} & IN & - & DINO & \phantom{1} 1.7 &  \phantom{1}0.1 &  \phantom{1}0.3 &  \phantom{1} - &  \phantom{1} - &  \phantom{1} - &&  \phantom{1} - &  \phantom{1} - &  \phantom{1} - &  \phantom{1} - &  \phantom{1} - &  \phantom{1} -\\
TokenCut~\cite{wang2022tokencut} & IN & - & DINO &  \phantom{1} - & \phantom{1} - &  \phantom{1} - &  \phantom{1} - &  \phantom{1} - &  \phantom{1} - &&  \phantom{1}\phantom{1}5.8 &  \phantom{1}2.8 &  \phantom{1}3.0 &  \phantom{1}4.8 & \phantom{1}1.9 &  \phantom{1}2.4 \\
\Ours (ours) & IN & MRCNN & DINO & \phantom{1}21.8	&	11.1	&	10.1	& 18.6	&	\phantom{1}9.0	&	\phantom{1}8.0 && \phantom{1}21.3	&	11.1	&	10.2	& 18.0	&	\phantom{1}8.9	&	\phantom{1}7.9 \\
\Ours (ours) & IN & Cascade & DINO & \phantom{1}22.4	&	12.5	&	11.9	& 19.6	&	10.0	&	\phantom{1}9.2 && \phantom{1}21.9	&	11.8	&	12.3 & 18.9	&	\phantom{1}9.7	& \phantom{1}9.2 \\
\multicolumn{2}{l}{\textit{vs. prev. SOTA}} &&& \plus{+12.7}	&	\plus{+9.3}	&	\plus{+7.8}	&	\plus{+9.9}	&	\plus{+6.6}	&	\plus{+4.9} && \plus{+12.3}	&	\plus{+8.7}	&	\plus{+8.1}	&	\plus{+9.5}	&	\plus{+6.4}	&	\plus{+4.9} \\
\hline
\end{tabular}
\end{center}
\vspace{-14pt}
\caption{\textbf{Unsupervised object detection and instance segmentation} on COCO \texttt{20K} and COCO \texttt{val2017}. We report the detection and segmentation metrics and note the \pretraining data (Pretrain), detectors, and backbone initialization (Init.).
Methods in the top half of the table train on extra unlabeled images from the downstream datasets, while zero-shot methods in the bottom half only train on \imnet.
Despite using an older detector, \Ours outperforms all prior works on all evaluation metrics.
$^*$: results obtained with the official code and checkpoint.
IN, Cascade, MRCNN, and FRCNN denote \imnet, Cascade Mask R-CNN, Mask R-CNN, and Faster R-CNN, respectively.
}
\label{tab:zero-shot-coco-coco20k}
\end{table*}
}
\definecolor{Gray}{gray}{0.9}
\newcommand{\tc}[1]{\textcolor{gray}{#1}}

\def\tabUVOAll#1{
\begin{table}[#1]
\vspace{-8pt}
\tablestyle{0.3pt}{1.0}
\small
\resizebox{0.48\textwidth}{!}{
\begin{tabular}{lHHcccccc}
Methods & & & \multicolumn{1}{c}{AP$_{50}^{\text{box}}$} & \multicolumn{1}{c}{AP$_{75}^{\text{box}}$} & \multicolumn{1}{c}{AP$^{\text{box}}$} & \multicolumn{1}{c}{AP$_{50}^{\text{mask}}$} & \multicolumn{1}{c}{AP$_{75}^{\text{mask}}$} & \multicolumn{1}{c}{AP$^{\text{mask}}$} \\ [.1em]
\shline
\multicolumn{1}{l}{\bf \textit{fully-supervised methods:}} &&&&&& \\
\tc{SOLO-v2 (w/ COCO)}\cite{wang2020solov2} & \tc{SOLO v2} & COCO$^{\text{X}}$ &- &	-&- & \phantom{1}\tc{38.0} & \tc{20.9} & \tc{21.4} \\
\tc{Mask R-CNN (w/ COCO)}\cite{he2017mask} & \tc{Mask R-CNN} & COCO$^{\text{X}}$ & -&	-& -&	\phantom{1}\tc{31.0} & \tc{14.2} & \tc{15.9} \\
\tc{SOLO-v2 (w/ LVIS)}\cite{wang2020solov2} & \tc{SOLO v2~\cite{wang2020solov2}} & LVIS & - &	-& -& \phantom{1}\tc{14.8} & \phantom{1}\tc{5.9} & \phantom{1}\tc{7.1} \\
\hline
\multicolumn{1}{l}{\bf \textit{unsupervised methods:}} &&&&&& \\
FreeSOLO$^*$~\cite{wang2022freesolo} & SOLO v2 & IN+COCO$^{\text{X+U}}$ & \phantom{1}{10.0}	&	\phantom{1} {1.8}	&	\phantom{1} {3.2} & \phantom{1} {9.5}	&	\phantom{1}{2.0}	&	\phantom{1}{3.3}	\\
\Ours (ours) & Cascade & IN & \phantom{1}31.7	&	\phantom{1}14.1	& \phantom{1}16.1	&  \phantom{1}22.8	&	\phantom{1}8.0	&	10.1	\\
\textit{vs. prev. SOTA} &&& \plus{+21.7}	&	\plus{+12.3}	&	\plus{+12.9} & \plus{+13.3}	&	\plus{+6.0}	&	\plus{+6.8} \\
\hline
\end{tabular}}
\vspace{-4pt}
\caption{Zero-shot unsupervised object detection and instance segmentation on the  \textbf{UVO \texttt{val} video benchmark}.
\Ours outperforms prior unsupervised methods and achieves better performance than the supervised SOLO-v2 model trained on the LVIS dataset.
$^*$: reproduced results with official code and checkpoint.
}
\label{tab:uvo}
\end{table}
}

\def\tabVOCAll#1{
\begin{table}[#1]
\vspace{-8pt}
\tablestyle{3.7pt}{1.0}
\small
\begin{center}
\begin{tabular}{p{2.2cm}HHHcccccc}
\multirow{1}{*}{Methods} & \multirow{1}{*}{Detector} &  \multirow{1}{*}{Init.} & \multirow{1}{*}{Pre-train} & \multicolumn{1}{c}{AP$_{50}$} & \multicolumn{1}{c}{AP$_{75}$} & \multicolumn{1}{c}{AP} & \multicolumn{1}{c}{AP$_{\text{S}}$} & \multicolumn{1}{c}{AP$_\text{M}$} & \multicolumn{1}{c}{AP$_{\text{L}}$} \\ [.1em]
\shline
rOSD \cite{vo2020toward} & RPN \cite{ren2015faster} & OBOW \cite{gidaris2021obow} & IN & \phantom{1}13.1 & - & \phantom{1} 4.3 & - & - & - \\
LOD \cite{vo2021large} & RPN \cite{ren2015faster} & Super. VGG \cite{simonyan2014very} & IN & \phantom{1}13.9 & - & \phantom{1} 4.5 & - & - & - \\
LOST \cite{simeoni2021localizing} & Faster \cite{ren2015faster} & DINO \cite{caron2021emerging} & IN+COCO$^{\text{X+U}}$ & \phantom{1}19.8 & - & \phantom{1} 6.7 & - & - & - \\
FreeSOLO$^*$ \cite{wang2022freesolo} & SOLOv2 \cite{wang2020solov2} & DenseCL \cite{wang2021dense} & IN+COCO$^{\text{X+U}}$ & \phantom{1}15.9 & \phantom{1} 3.6 & \phantom{1} 5.9 & \phantom{1}0.0 & \phantom{1}2.0 & \phantom{1} 9.3 \\
\hline
\Ours (ours) & Cascade \cite{cai2018cascade} & DINO \cite{caron2021emerging} & IN & \phantom{1}36.9	&	\phantom{1}19.2	&	\phantom{1}20.2	&	\phantom{1}1.3	&	\phantom{1}6.5	&	\phantom{1}32.2 \\
\textit{vs. prev. SOTA} &&&& \plus{+17.1}	&	\plus{+15.6}	&	\plus{+13.5}	&	\plus{+1.3}	&	\plus{+4.5}	&	\plus{+22.9} \\
\hline
\end{tabular}
\end{center}
\vspace{-14pt}
\caption{Zero-shot unsupervised object detection on \textbf{VOC}. $^*$: reproduced results with official code and checkpoint.}
\label{tab:voc}
\end{table}
}

\def\tabAllDatasets#1{
\begin{table*}[#1]
\tablestyle{1pt}{1.0}
\small
\begin{center}
\begin{tabular}{p{2.2cm}|p{1.2cm}p{1.2cm}p{1.2cm}p{1.2cm}p{1.2cm}p{1.2cm}p{1.2cm}p{1.2cm}p{1.2cm}p{1.2cm}p{1.2cm}}
\multirow{1}{*}{Datasets} & \ROTATED{COCO} & \ROTATED{COCO20K} & \ROTATED{VOC} & \ROTATED{LVIS} &\ROTATED{UVO} & \ROTATED{Clipart} & \ROTATED{Comic}  & \ROTATED{Watercolor} & \ROTATED{KITTI} & \ROTATED{Objects365} & \ROTATED{OpenImages} \\ [3.5em]
\shline
Baseline & 12.2 & 9.7 & 15.9 & 3.8 & 10.0 & 7.9 & 9.9 & 6.7 & 7.7 & 8.1 & 9.9 \\
\hline
\Ours & 22.7	&	23.2	&	38.9	&	8.7	&	28.3	&	19.4 & 28.4 & 31.5 & 10.4 & 20.2 & 16.1 \\
$\Delta$ & \plus{+10.5}	&	\plus{+12.5}	&	\plus{+23.0}	&	\plus{+18.3}	&	\plus{+11.5} &	\plus{+11.5} & \plus{+18.5} & \plus{+25.2} & \plus{+2.7} & \plus{+12.1} & \plus{+6.2} \\
\end{tabular}
\end{center}
\vspace{-4mm}
\caption{Zero-shot unsupervised multi-object discovery on various benchmarks, \ie optimized on \imnet and inferred on various datasets. Evaluation metric: AP$^{\text{box}}_{50}$. Baseline method is FreeSOLO \cite{wang2022freesolo} and is tested with its official code and released checkpoint.}
\label{tab:all-datasets}
\end{table*}
}

\def\tabFineTuneMRCNN#1{
\begin{figure}[#1]
  \centering
  \includegraphics[width=1.0\linewidth]{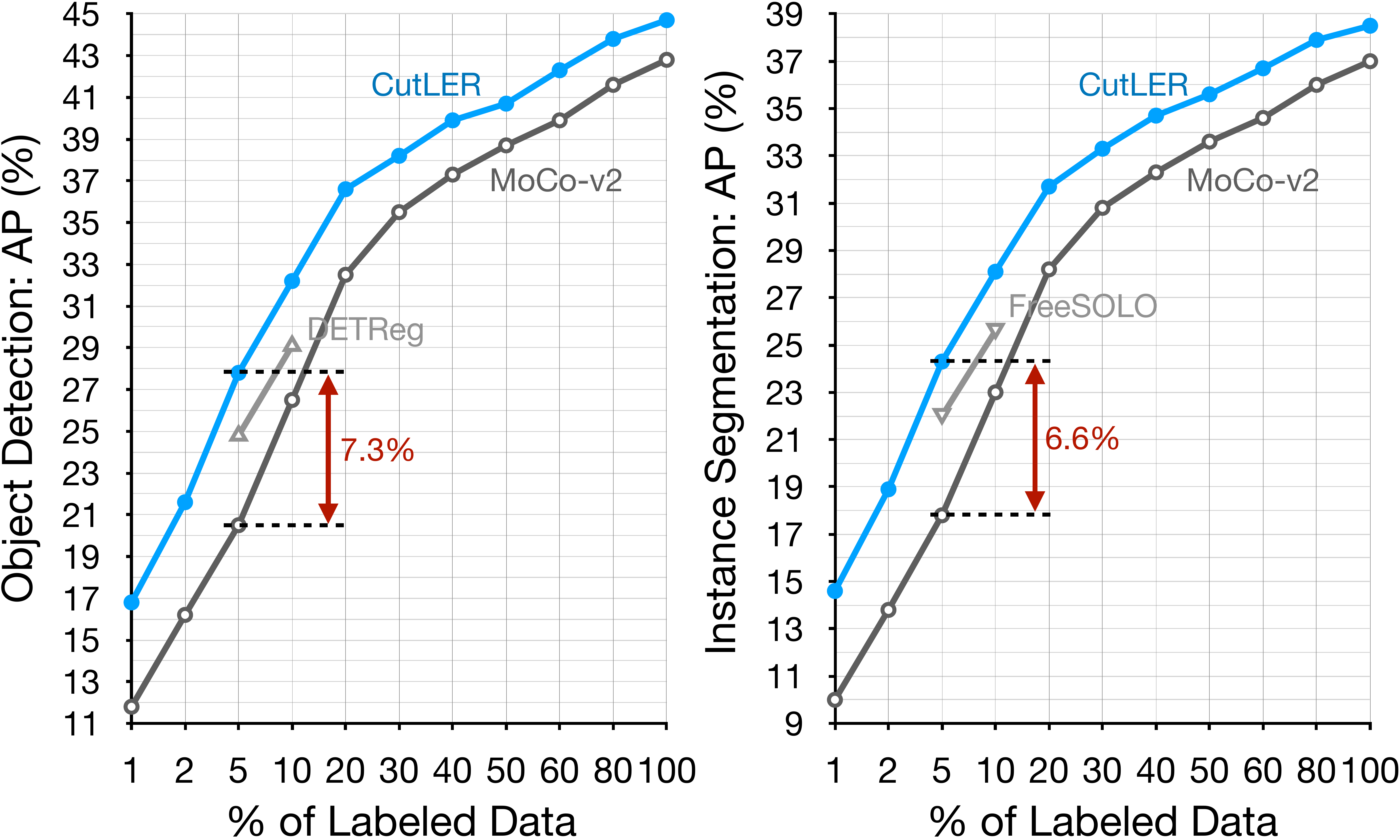}
  \vspace{-10pt}
  \caption{\textbf{Finetuning \Ours for low-shot and fully supervised detection and instance segmentation.}
  We fine-tune a Cascade Mask R-CNN model initialized with \Ours or MoCo-v2 on varying amounts of labeled data on the COCO dataset.
  We use the same schedule as the self-supervised pretrained MoCo-v2 counterpart and report the detection and instance segmentation performance.
  \Ours consistently outperforms the MoCo-v2 baseline: in the low-shot setting with 1\% labels and the fully supervised setting using 100\% labels.
  \Ours also outperforms FreeSOLO~\cite{wang2022freesolo} and DETReg~\cite{bar2022detreg} on this benchmark despite using an older detection architecture. Results with Mask R-CNN are in the appendix.
  }
  \label{fig:fine-tune-coco}
\end{figure}
}

\def\tabVSSOTA#1{
\begin{figure}[#1]
  \centering
  \includegraphics[width=1.0\linewidth]{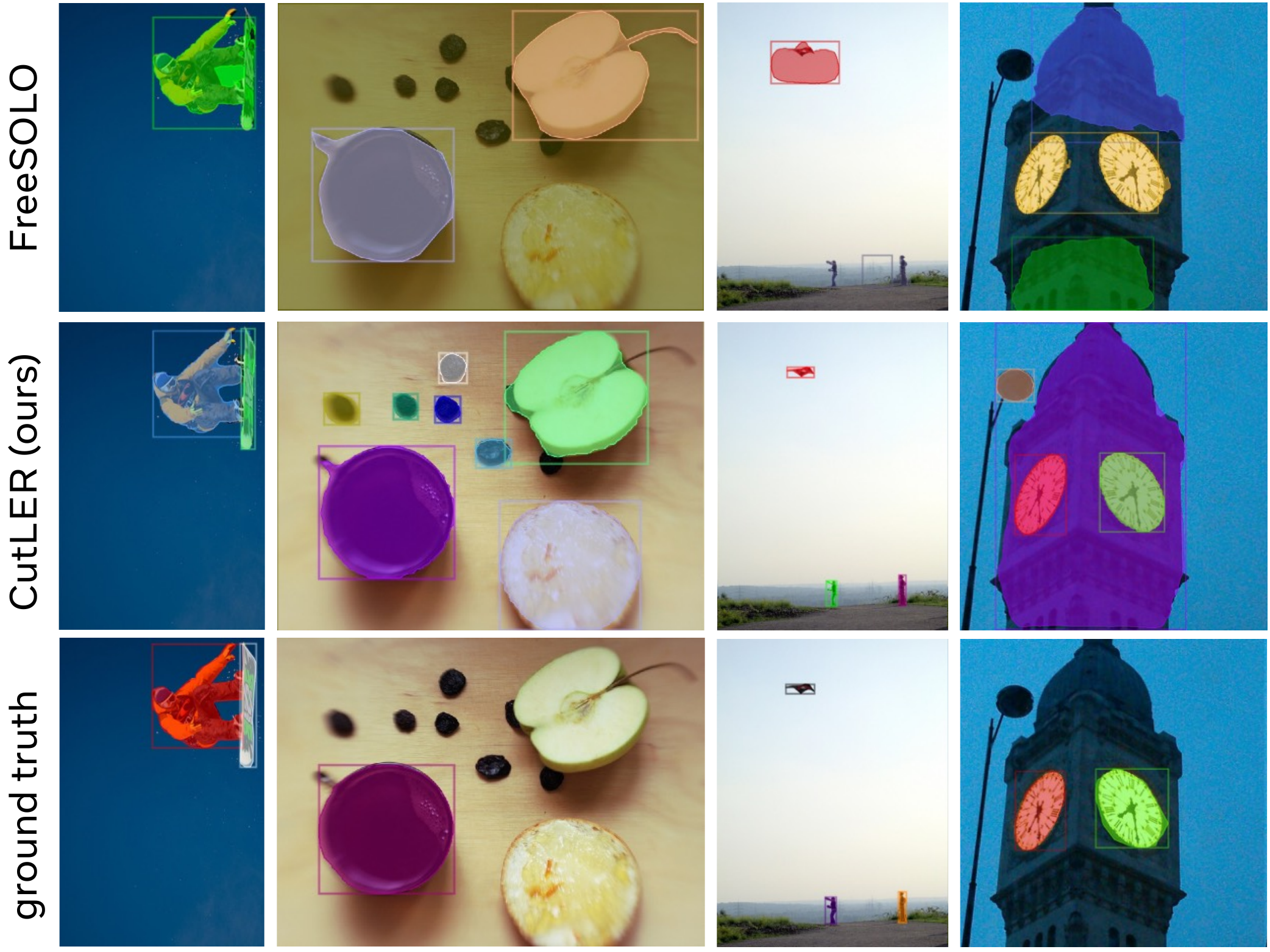}\vspace{-4pt}
  \caption{
  \textit{Compared to the previous state-of-the-art} \cite{wang2022freesolo}, our \Ours can better discriminate instances (\eg person and skis in col.~1), discover more objects (\eg apple and raisins in col.~2), and produce higher quality segmentation masks even for small objects (\eg kite in col.~3); \textit{compared to human annotations}, \Ours can locate novel instances that are overlooked by human annotators, such as the streetlight and clock tower in col.~4.
  \textbf{Qualitative comparisons} between previous SOTA methods (row 1) and our \Ours (row 2) on COCO, as well as ground truth annotations by human annotators (row 3), are visualized.
  }
  \label{fig:vs-sota}
\end{figure}
}


\section{Experiments}
\label{experiments}

We evaluate \Ours on various detection and segmentation benchmarks.
In~\cref{sec:exp-unuspervised-zs-seg}, we show that \Ours can discover objects without any supervision on completely unseen images.
Despite being evaluated in a zero-shot manner on eleven benchmarks, \Ours outperforms prior methods that use in-domain training data.
\cref{sec:exp-finetune-eval} shows that finetuning \Ours further improves detection performance, outperforming prior work like MoCo-V2 and FreeSOLO.
 
\tabCOCOplusTwntyK{!t}
\subsection{Unsupervised Zero-shot Evaluations}
\label{sec:exp-unuspervised-zs-seg}
We conduct extensive experiments on eleven different datasets, covering various object categories, image styles, video frames, resolutions, camera angles, \etc to verify the effectiveness of \Ours as a universal unsupervised object detection and segmentation method.
We describe the different datasets used for zero-shot evaluation in detail in~\cref{appendix:zero-shot-dataset-details}.
\Ours is trained solely using images from ImageNet and evaluated in a zero-shot manner on all downstream datasets without finetuning on any labels or data.

\par \noindent \textbf{Evaluating unsupervised object detectors} poses two unique challenges.
\textit{First}, since the model is trained without any notion of semantic classes, it cannot be evaluated using the class-aware detection setup.
Thus, like prior work~\cite{wang2020solov2,bar2022detreg,simeoni2021localizing} we use the class-agnostic detection evaluation.
\textit{Second}, object detection datasets often only annotate a subset of the objects in the images.
For example, while COCO and LVIS use the same images, COCO only labels 80 object classes, and LVIS labels 1203 object classes.
In this partially labeled setup, Average Recall (AR) is a valuable metric for unsupervised detection as it does not penalize the models for detecting novel objects unlabeled in the dataset.
Thus, we additionally report AR for all datasets.

\par \noindent \textbf{Zero-shot detection on 11 benchmarks.} We evaluate \Ours on a variety of datasets and report the detection performance using AP$^\text{box}_{50}$ and AR$^\text{box}_{100}$ metrics in~\cref{fig:teaser} and~\cref{tab:zero-shot-all-datasets}.
\Ours uses a \emph{smaller} model size and \emph{less} training data than prior work.
Compared to the previous SOTA approach, FreeSOLO~\cite{wang2022freesolo} with a backbone of ResNet101, \Ours, with the smaller ResNet50 backbone, significantly outperforms it in each of these benchmarks spanning various image distributions, more than doubling performance on 10 of them. Also note that, FreeSOLO requires FreeMask pre-training using approximately 1.3M ImageNet images and model fine-tuning using additional data in test benchmarks.

We observe that on different domains, \eg watercolor or frames from videos (UVO dataset), \Ours improves performance by over $4\times$ and $2\times$, respectively.
\cref{fig:teaser} shows some qualitative examples of \Ours's predictions.

\tabVSSOTA{!t}
\par \noindent \textbf{Detailed comparisons on COCO20K and COCO}.
\cref{tab:zero-shot-coco-coco20k} presents detailed detection and segmentation evaluations (also referred to as `multi-object' discovery) on two popular benchmarks: COCO \texttt{val2017}~\cite{lin2014microsoft} and COCO \texttt{20K}, which contains a subset of 20K images of COCO~\cite{wang2022freesolo,simeoni2021localizing}.
\Ours consistently surpasses prior works by a large margin (often gets 2$\app$3$\times$ higher AP) on both the segmentation and detection tasks.
Although \Ours is not trained on any images from COCO, it surpasses existing methods trained on COCO by more than 10\% in terms of AP$^\text{mask}_{50}$ and AP$^\text{box}_{50}$.

\cref{fig:vs-sota} shows the qualitative comparisons between~\cite{wang2022freesolo} and our \Ours on COCO \texttt{val2017}, along with human annotations.
Surprisingly, \textit{\Ours can often detect novel instances that human annotators miss}. 

We present detailed comparisons on COCO \texttt{20K}, COCO \texttt{val2017} and LVIS~\cite{gupta2019lvis} benchmarks in~\cref{appendix:zero-shot-additional}.

\tabVOCAll{t!}

\par \noindent \textbf{Detailed comparisons on UVO and VOC}.
For a comprehensive comparison with existing unsupervised multi-object detection methods, we report the results for UVO \texttt{val}~\cite{wang2021unidentified} and VOC \texttt{trainval07}~\cite{everingham2010pascal}.
\cref{tab:voc} shows that \Ours yields significant performance gains over previous SOTA, obtaining over 3$\times$ higher AP, with the most considerable improvement coming from AP$_{L}$.
On UVO, \cref{tab:uvo} shows that \Ours more than quadruples the AP of previous SOTA and almost triples the AP$^\text{box}_{50}$. \textit{Our AP$^\text{mask}_{50}$ is even 4.8\% higher than the fully-supervised SOLOv2~\cite{wang2020solov2} trained on LVIS with 100\% annotations}, significantly narrowing the gap between supervised and unsupervised learning.
\tabUVOAll{t!}

\subsection{Label-Efficient and Fully-Supervised Learning} 
\label{sec:exp-finetune-eval}

We now evaluate \Ours as a pretraining method for training object detection and instance segmentation models.
While \Ours can discover objects without any supervision, finetuning it on a target dataset aligns the model output to the same set of objects labeled in the dataset.

\tabFineTuneMRCNN{t!}

\par \noindent \textbf{Setup.}
We use \Ours to initialize a standard Cascade Mask R-CNN~\cite{cai2018cascade} detector with a ResNet50~\cite{he2016deep}.
Prior work uses more advanced detectors, SOLOv2~\cite{wang2020solov2} used in~\cite{wang2022freesolo} and DDETR \cite{zhu2020deformable} used in~\cite{bar2022detreg}, that perform better.
However, we choose Cascade Mask R-CNN for its simplicity and show in~\cref{sec:ablation} that \Ours's performance improves with stronger detectors.
We train the detector on the COCO~\cite{lin2014microsoft} dataset using the bounding box and instance mask labels.
To evaluate label efficiency, we subsample the training set to create subsets with varying proportions of labeled images.
We train the detector, initialized with \Ours, on each of these subsets.
As a baseline, we follow the settings from MoCo-v2~\cite{chen2020improved} and train the same detection architecture initialized with a MoCo-v2 ResNet50 model, given its strong performance on object detection tasks.
Both MoCo-v2 and our models are trained for the $1\times$ schedule using Detectron2~\cite{wu2019detectron2}, except for extremely low-shot settings with 1\% or 2\% labels. Following previous works~\cite{wang2022freesolo}, when training with 1\% or 2\% labels, we train both MoCo-v2 and our model for 3,600 iterations with a batch size of 16.

\par \noindent \textbf{Results.}
\cref{fig:fine-tune-coco} shows the results of fine-tuning the detector on different subsets of COCO. When tested with low-shot settings, \eg, 2\% and 5\% labeled data, our approach achieves 5.4\% and 7.3\% higher AP$^{\text{box}}$ than the MoCo-v2 baseline, respectively.
Even when training with full annotations, \Ours still consistently gives more than 2\% improvements, outperforming MoCo-v2 for both object detection and segmentation.
More impressively, \Ours outperforms prior SOTA methods - FreeSOLO~\cite{wang2022freesolo} and DETReg~\cite{bar2022detreg} despite using an older detection architecture.
\def\tabModels#1{
\begin{table}[#1]
\tablestyle{1pt}{1.0}
\small
\begin{center}
\begin{tabular}{lcacc}
& Mask R-CNN & Cascade Mask R-CNN & ViTDet \\ [.1em]
\shline
\begin{tabular}[c]{@{}p{24pt}p{3pt}p{24pt}} AP$^{\text{box}}_{50}$ & / & AP$^{\text{box}}$ \end{tabular}
& 20.3 / 10.6 & 20.8 / 11.5 & 21.5 / 11.8 \\
\begin{tabular}[c]{@{}p{24pt}p{3pt}p{24pt}} AP$^{\text{mask}}_{50}$ & / & AP$^{\text{mask}}$ \end{tabular} & 17.2 / \phantom{1}8.5 & 17.7 / \phantom{1}8.8 & 18.0 / \phantom{1}9.0 \\
\hline
\end{tabular}
\end{center}
\vspace{-14pt}
\caption{\textbf{\Ours with different detection architectures}. We report results on COCO and observe that \Ours is agnostic to the detection architecture and improves performance using stronger detection architectures such as ViTDet with a backbone of ViT-B.}
\label{tab:abl-models}
\end{table}
}

\def\tabNumRounds#1{
\begin{table}[#1]
\tablestyle{3.5pt}{1.0}
\small
\begin{center}
\begin{tabular}{lccclccc}
    & \multicolumn{3}{c}{UVO} && \multicolumn{3}{c}{COCO} \\
    \cline{2-4} \cline{6-8}
    & \multicolumn{1}{c}{AP$_{50}^{\text{mask}}$} & \multicolumn{1}{c}{AP$^{\text{mask}}$} & \multicolumn{1}{c}{AP$_{75}^{\text{mask}}$} && \multicolumn{1}{c}{AP$_{50}^{\text{mask}}$} & \multicolumn{1}{c}{AP$^{\text{mask}}$} & \multicolumn{1}{c}{AP$_{75}^{\text{mask}}$} \\ [.1em]
\shline
1 round & 20.6 & \phantom{1}9.0 & 7.0 && 17.7 & 8.8 & 8.0 \\
2 rounds & 22.2 & \phantom{1}9.6 & 7.5 && 18.5 & 9.5 & 8.8 \\
\rowcolor{Gray} 3 rounds & 22.8 & 10.1 & 8.0 && 18.9 & 9.7 & 9.2 \\
4 rounds & 22.8 & 10.2 & 8.2 && 18.9 & 9.8 & 9.3 \\
\hline
\end{tabular}
\end{center}
\vspace{-14pt}
\caption{\textbf{Number of self-training rounds} used in \Ours. We find that 3 rounds of self-training are sufficient. Self-training provides larger gains for the densely labeled UVO dataset.
}
\label{tab:abl-self-training}
\end{table}
}

\def\tabComponents#1{
\begin{table}[#1]
\tablestyle{5pt}{1.0}
\small
\begin{center}
\begin{tabular}{lcclcc}
\multirow{2}{*}{Methods} & \multicolumn{2}{c}{UVO} && \multicolumn{2}{c}{COCO} \\
\cline{2-3} \cline{5-6}
 & \multicolumn{1}{c}{AP$_{50}^{\text{mask}}$} & \multicolumn{1}{c}{AP$^{\text{mask}}$} && \multicolumn{1}{c}{AP$_{50}^{\text{mask}}$} & \multicolumn{1}{c}{AP$^{\text{mask}}$} \\ [.1em]
\shline
TokenCut \cite{wang2022tokencut} & - & - && 4.9 & 2.0 \\
\hline
Base & 14.6 & 5.4 && 13.5 & 5.7 \\
+ MaskCut & 19.3 & 8.1 && 15.8 & 7.7 \\
+ \droploss & 20.9 & 9.0 && 16.6 & 8.2 \\
+ copy-paste~\cite{ghiasi2021simple,dwibedi2017cut} & 21.5 & 9.9 && 17.7 & 8.8 \\
\rowcolor{Gray} + self-train (\Ours) & 22.8 & 10.1 && 18.9 & 9.7 \\
\hline
\end{tabular}
\end{center}
\vspace{-14pt}
\caption{Ablation study on the contribution of each component. Results reported on COCO and video segmentation dataset UVO.}
\label{tab:abl-components}
\end{table}
}

\def\tabTokenCut#1{
\begin{table}[#1]
\tablestyle{0.4pt}{1.0}
\small
\begin{tabular}{lcHcHHclcHcHHc}
 Methods & \multicolumn{1}{c}{AP$_{50}^{\text{box}}$} & \multicolumn{1}{H}{AP$_{75}^{\text{box}}$} & \multicolumn{1}{c}{AP$^{\text{box}}$} & \multicolumn{1}{H}{AR$_1^{\text{box}}$} & \multicolumn{1}{H}{AR$_{10}^{\text{box}}$} & \multicolumn{1}{c}{AR$_{100}^{\text{box}}$} && \multicolumn{1}{c}{AP$_{50}^{\text{mask}}$} & \multicolumn{1}{H}{AP$_{75}^{\text{mask}}$} & \multicolumn{1}{c}{AP$^{\text{mask}}$} & \multicolumn{1}{H}{AR$_1^{\text{mask}}$} & \multicolumn{1}{H}{AR$_{10}^{\text{mask}}$} & \multicolumn{1}{c}{AR$_{100}^{\text{mask}}$} \\ [.1em]
\shline
TokenCut (1 eigenvec.) & \phantom{1}5.2 & \phantom{1}2.7 & \phantom{1}2.6 & \phantom{1}5.0 & \phantom{1}5.0 & \phantom{1}5.0 && \phantom{1}4.9 & 2.3 & 2.0 & 4.4 & 4.4 & \phantom{1}4.4 \\
TokenCut (3 eigenvec.) & \phantom{1}4.7 & \phantom{1}2.1 & \phantom{1}1.7 & \phantom{1}5.4 & \phantom{1}8.1 & \phantom{1}8.1 && \phantom{1}3.6 & 1.6 & 1.2 & 4.8 & 6.9 & \phantom{1}6.9 \\
\hline 
MaskCut ($t=3$) & \phantom{1}6.0 & 2.3 & \phantom{1}2.9 & \phantom{1}5.4 & \phantom{1}8.1 & \phantom{1}8.1 && \phantom{1}4.9 & 1.7 & 2.2 & 4.7 & 6.9 & \phantom{1}6.9 \\
\rowcolor{Gray} \Ours & 21.9 & 11.8 & 12.3 & 6.8 & 19.6 & 32.7 && 18.9 & 9.2 & 9.7 & 5.8 & 16.5 & 27.1 \\
\hline
\end{tabular}
\vspace{-4pt}
\caption{\Ours achieves much higher results even when compared to a modified TokenCut that can produce more than one mask per image. Compared to TokenCut, MaskCut gets a higher recall without reducing precision. We report results on COCO.}
\label{tab:abl-tokencut}
\end{table}
}

\def\tabDatasets#1{
\begin{table}[#1]
\tablestyle{1.6pt}{1.0}
\small
\begin{center}
\begin{tabular}{llccclcccc}
Pre-train & \Ours & \multicolumn{1}{c}{AP$_{50}^{\text{box}}$} & \multicolumn{1}{c}{AP$_{75}^{\text{box}}$} & \multicolumn{1}{c}{AP$^{\text{box}}$} && \multicolumn{1}{c}{AP$_{50}^{\text{mask}}$} & \multicolumn{1}{c}{AP$_{75}^{\text{mask}}$} & \multicolumn{1}{c}{AP$^{\text{mask}}$}\\ [.1em]
\shline
\rowcolor{Gray} IN1K & IN1K & 20.8 & 10.8 & 11.5 && 17.7 & 8.0 & 8.8 \\
YFCC1M & YFCC1M & 19.4 & 10.4 & 10.9 && 16.3 & 7.4 & 8.1 \\
IN1K & YFCC1M & 14.9 & \phantom{1}7.6 & \phantom{1}8.2 && 12.1 & 5.4 & 5.9 \\
YFCC1M & IN1K & 14.8 & \phantom{1}7.2 & \phantom{1}8.0 && 11.8 & 5.2 & 5.8 \\
\hline
\end{tabular}
\end{center}
\vspace{-14pt}
\caption{\textbf{Impact of datasets} used to pre-train DINO and train \Ours. 
\Ours's detection performance is similar when pretraining both DINO and \Ours with the same dataset: the object-centric \imnet dataset or the non-object-centric YFCC dataset.
}
\label{tab:abl-datasets}
\end{table}
}

\def\tabAblationsMaskCutDropLoss#1{
\begin{table*}[#1]
	\centering
	\subfloat[
	\textbf{Image size}.
	\label{tab:ablate_image_size}
	]{
		\centering
		\begin{minipage}{0.25\linewidth}{\begin{center}
					\tablestyle{1.5pt}{1.2}
					\begin{tabular}{cccac}
                        Size $\rightarrow$ & 240 & 360 & 480 & 640 \\ [.1em]
                        \shline
						AP$^{\text{mask}}_{50}$ & 15.1 & 16.6 & 17.7 & 17.9 \\
                        \hline
                    \end{tabular}
		\end{center}}\end{minipage}
	}
	\subfloat[
	\textbf{$\tau^{\text{ncut}}$ for \maskcut}.
	\label{tab:ablate_tau_ncut}
	]{
		\begin{minipage}{0.25\linewidth}{\begin{center}
					\tablestyle{1.5pt}{1.2}
					\begin{tabular}{cccacc}
                        $\tau^{\text{ncut}}$ $\rightarrow$ & 0 & 0.1 & 0.15 & 0.2 & 0.3 \\ [.1em]
                        \shline
						AP$^{\text{mask}}_{50}$ & 17.1 & 17.5 & 17.7 & 17.6 & 17.5 \\
                        \hline
                    \end{tabular}
		\end{center}}\end{minipage}
	}
	\subfloat[
	\textbf{\# masks per image}.
	\label{tab:ablate_num_masks}
	]{
		\begin{minipage}{0.25\linewidth}{\begin{center}
					\tablestyle{1.5pt}{1.2}
					\begin{tabular}{ccac}
                        $N$ $\rightarrow$ & 2 & 3 & 4 \\ [.1em]
                        \shline
						AP$^{\text{mask}}_{50}$ & 16.9 & 17.7 & 17.7 \\
                        \hline
                    \end{tabular}
		\end{center}}\end{minipage}
	}
    \subfloat[
	\textbf{$\tau^{\text{IoU}}$ for \droploss}.
	\label{tab:ablate_tau_droploss}
	]{
		\begin{minipage}{0.22\linewidth}{\begin{center}
					\tablestyle{1.5pt}{1.2}
					\begin{tabular}{ccacc}
                        $\tau^{\text{IoU}}$ $\rightarrow$ & 0 & 0.01 & 0.1 & 0.2 \\ [.1em]
                        \shline
						AP$^{\text{mask}}_{50}$ & 17.4 & 17.7 & 14.4 & 12.7 \\
                        \hline
                    \end{tabular}
		\end{center}}\end{minipage}
	}\vspace{-6pt}
	\caption{\textbf{Ablations for \maskcut and \droploss} used for training \Ours. We report  \Ours's detection and instance segmentation performance on COCO \texttt{val2017}, without adding the self-training stage. \textbf{(a)} We vary the size of the image used for \maskcut. \textbf{(b)} We vary the threshold $\tau^{\text{ncut}}$ in \maskcut, which controls the sparsity of the affinity matrix used for Normalized Cuts. \textbf{(c)} We vary the number of masks extracted using \maskcut and train different \Ours models. \textbf{(d)} We vary $\tau^{\text{IoU}}$ in \droploss, \ie, the maximum overlap between the predicted regions and the ground truth beyond which the loss for the predicted regions is ignored. Default settings are highlighted in \colorrowtext{}.}
	\label{tab:ablate_maskcut_droploss}
\end{table*}
}

\def\tabMultiRounds#1{
\begin{figure}[#1]
  \vspace{-4pt}
  \centering
  \includegraphics[width=0.96\linewidth]{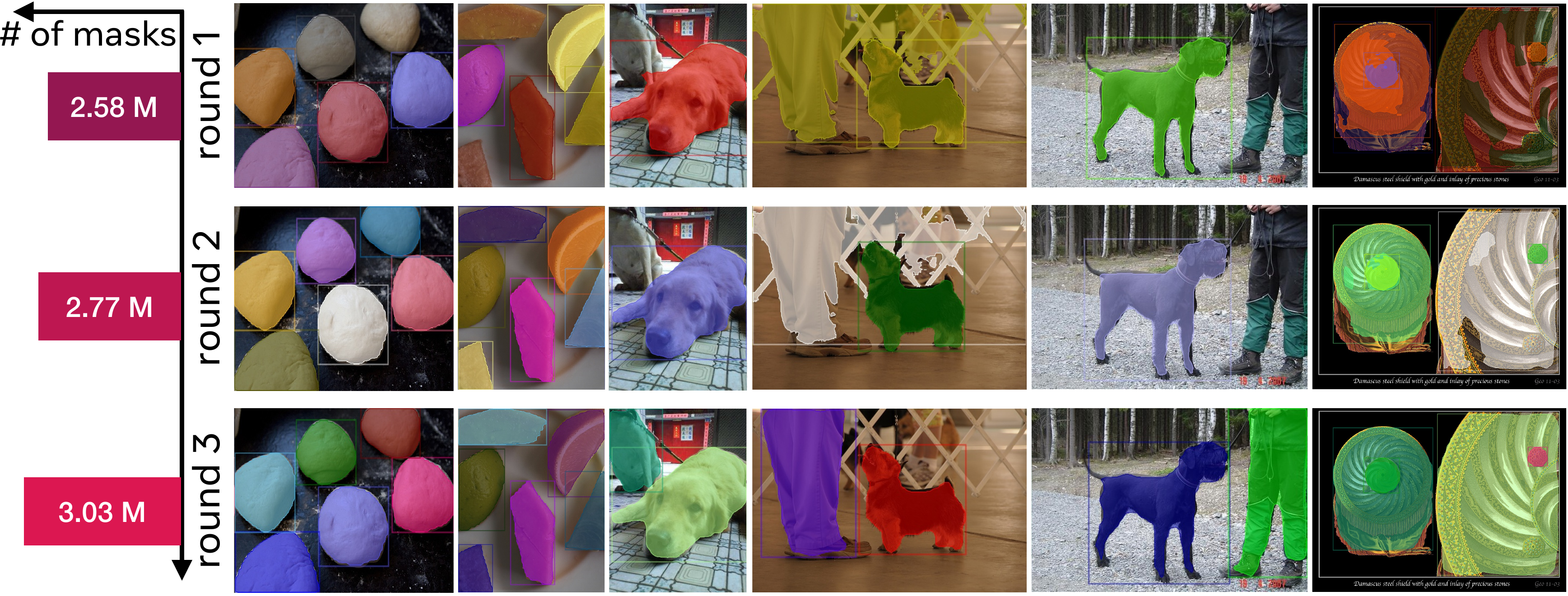}\vspace{-4pt}
  \caption{
  Multiple rounds of self-training can improve the pseudo-masks in terms of quality and quantity. 
  We show \textbf{qualitative visualizations and the number of pseudo-masks} for all three rounds.
  }
  \label{fig:multi-rounds}
\end{figure}
}

\section{Ablations}
\label{sec:ablation}
We analyze the design decisions in \Ours.
We use similar settings to~\cref{experiments} and train \Ours only on ImageNet.
We use the Cascade Mask R-CNN detection architecture and evaluate our model primarily on the COCO and UVO unsupervised detection benchmarks.
All ablation studies are conducted without self-training unless otherwise noted.

\tabComponents{t!}
\tabTokenCut{t!}

\tabAblationsMaskCutDropLoss{t!}

\paragraph{Importance of each component.}
We analyze the main components of \Ours and report their relative contribution in~\cref{tab:abl-components}.
We report results on the popular COCO~\cite{lin2014microsoft} dataset and a densely annotated video instance segmentation dataset UVO~\cite{wang2021unidentified}.
We also report the performance of running TokenCut~\cite{wang2022tokencut} on the COCO dataset.
Next, we use TokenCut's official codes to generate masks on ImageNet and use them for training a Cascade Mask R-CNN~\cite{cai2018cascade}.
This base model provides substantial gains over just using TokenCut on COCO. We add each of our proposed components to this strong base model.
Using MaskCut increases AP$_{50}^\text{mask}$ and AP$^\text{mask}$ by 4.7\% and 2.7\%, respectively. Also, the improvements to AP$_{50}^\text{mask}$ is larger for densely annotated dataset UVO, \ie 4.7\% vs. 2.7\%.
These results prove that \maskcut's ability to segment multiple instances per image is vital for densely annotated datasets.
Adding \droploss brings another 1.6\% and 0.9\% improvements to AP$_{50}^\text{mask}$ for UVO and COCO, respectively.
Multi-round of self-training increases the quantity and quality of pseudo-masks, leading to 1.3\% improvements.
These results show that each simple proposed component is critical for strong performance.

\paragraph{Comparison with TokenCut.}
TokenCut~\cite{wang2022tokencut} is also a zero-shot segmentation method.
However, it only segments a single instance per image, as discussed in~\cref{sec:preliminaries}.
In order to generate more than one segmentation mask per image, we use a modified TokenCut by using more of the smaller eigenvectors and combining all produced masks.
~\cref{tab:abl-tokencut} shows the object detection performance on COCO's validation set for vanilla TokenCut, our modified TokenCut and \Ours.
Although using more eigenvectors increases the recall AR$_{100}^\text{box}$, it significantly reduces the precision AP$^\text{box}$. 
\Ours not only improves the average recall AR$_{100}^\text{box}$ by 4$\times$ but also surpasses TokenCut's average precision AP$^\text{box}$ by 4.8$\times$, \ie 480\% relative improvements.

\paragraph{Design choices in \maskcut and \droploss} and their impact on the final localization performance is presented in~\cref{tab:ablate_maskcut_droploss}.
We first study the effect of the image size used by \maskcut for generating the initial masks.
As expected, \cref{tab:ablate_image_size} shows that \maskcut benefits from using higher resolution images presumably as it provides a higher resolution similarity between pixels.
We pick a resolution of 480px for a better trade-off between the speed of \maskcut and its performance.
In~\cref{tab:ablate_tau_ncut}, we study the effect of the threshold used in \maskcut for producing a binary $W$ matrix (\cref{sec:masked-ncuts}).
Overall, \Ours seems to be robust to the threshold values.
We understand the impact of the number of masks per image generated by \maskcut in~\cref{tab:ablate_num_masks}.
Increasing the number improves the performance of the resulting \Ours models.
This shows that \maskcut generates high-quality masks that directly impact the overall performance.
Finally, in~\cref{tab:ablate_tau_droploss}, we vary the IOU threshold used for \droploss.
With a high threshold, we ignore the loss for a higher number of predicted regions while encouraging the model to explore.
$0.01$ works best for the trade-off between exploration and detection performance.

\tabNumRounds{t!}
\tabMultiRounds{t!}

\paragraph{Self-training} and its impact on the final performance is analyzed in~\cref{tab:abl-self-training}.
Self-training consistently improves performance across the UVO and COCO benchmarks and all metrics.
UVO, which has dense object annotations, benefits more from the multi-round of self-training. By default, \Ours uses 3 rounds of self-training.
\cref{fig:multi-rounds} shows qualitative examples of how self-training improves both the quality of predictions and the number of objects predicted.

\tabModels{t!}
\paragraph{Generalization to different detection architectures.}
We use different detector architectures for training \Ours and measure their performance in~\cref{tab:abl-models}.
We observe that \Ours works with various architectures, and its performance is improved with stronger architectures.

\tabDatasets{t!}
\paragraph{Impact of the pretraining dataset.}
We now study the impact of the dataset used for 1) pretraining the self-supervised DINO model and 2) training the \Ours model. 
The commonly used \imnet dataset has a well-known object-centric bias~\cite{deng2009imagenet} which may affect the unsupervised detection performance.
Thus, we also use YFCC~\cite{thomee2016yfcc100m}, a non-object-centric dataset.
We control for the number of images in both \imnet and YFCC for a fair comparison and use them for training DINO and \Ours.
As~\cref{tab:abl-datasets} shows, \Ours's performance on COCO is robust to the choice of object-centric or non-object-centric datasets as long as the same dataset is used to train DINO and \Ours.
This shows the generalization of \Ours to different data distributions.
However, training DINO and \Ours with different data leads to worse performance, suggesting the importance of using the same image distribution for learning both DINO and \Ours models.
\section{Summary}
Object localization is a fundamental task in computer vision.
In this paper, we have shown that a simple yet effective cut-and-learn approach can achieve extraordinary performance on challenging object detection and instance segmentation tasks without needing to train with human annotations. As a zero-shot unsupervised detector, \Ours, trained solely on \imnet, outperforms the detection performance of previous works by over 2.7$\times$ on 11 benchmarks across various domains.

{\small
\bibliographystyle{ieee_fullname}
\bibliography{egbib}
}

\def\tabSelectiveSearch#1{
\begin{figure}[#1]
  \centering
  \begin{tabular}{c}
  \includegraphics[width=0.95\linewidth]{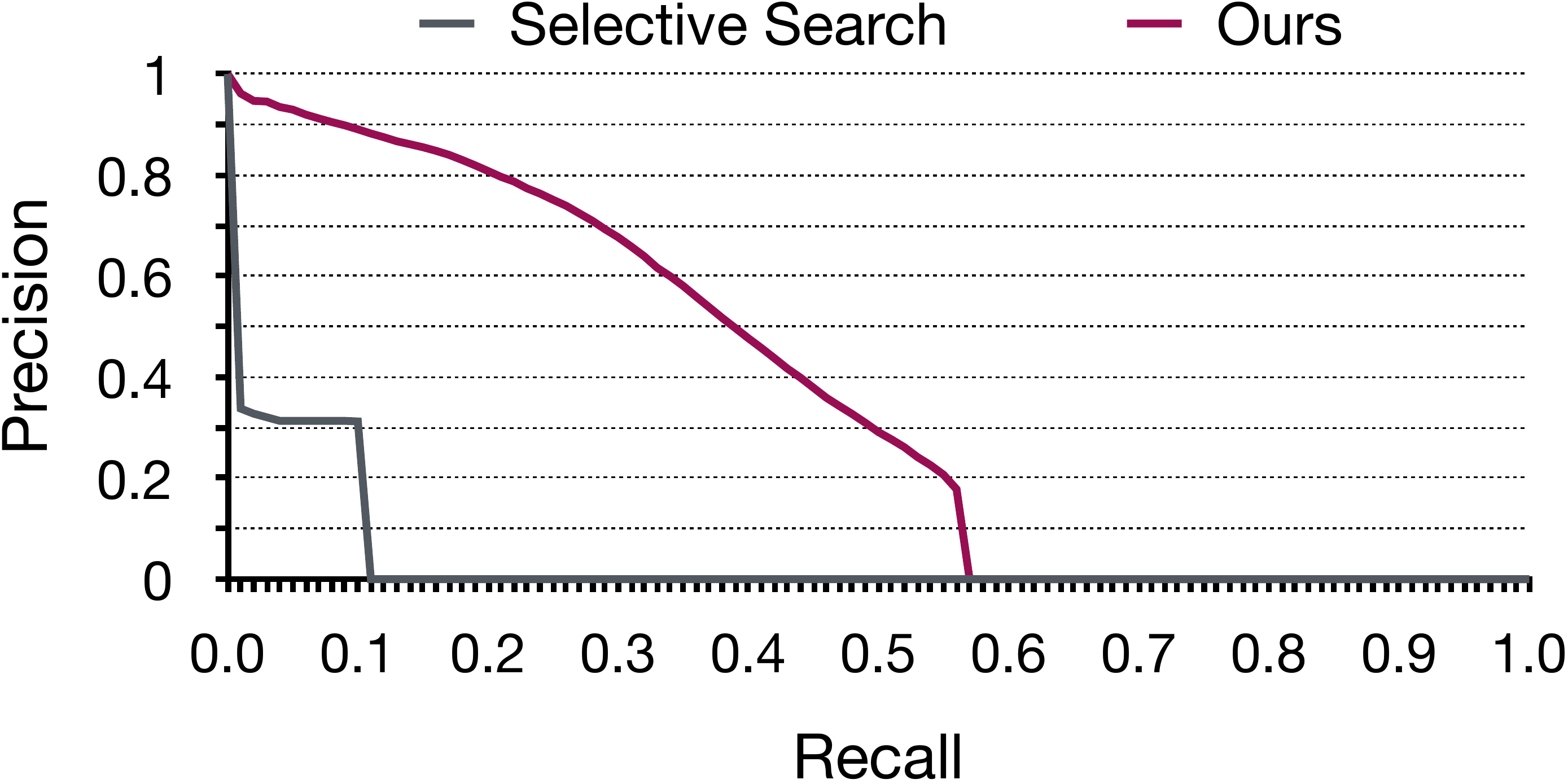}
  \end{tabular}
  \caption{Precision-recall curve for comparing selective search and CutLER on VOC07 \texttt{trainval}.}
  \label{fig:selective-search}
\end{figure}
}
\def\tabLVISAll#1{
\begin{table*}[#1]
\tablestyle{1pt}{1.0}
\small
\begin{center}
\begin{tabular}{p{1.6cm}|p{1.3cm}p{1.3cm}p{1.8cm}|p{0.85cm}p{0.85cm}p{0.85cm}p{0.85cm}p{0.85cm}p{0.85cm}|p{0.85cm}p{0.85cm}p{0.85cm}p{0.85cm}p{0.85cm}p{0.85cm}}
\multirow{2}{*}{Methods} & \multirow{2}{*}{Detector} &  \multirow{2}{*}{Init.} & \multirow{2}{*}{Pre-train} & \multicolumn{6}{c|}{Object Detection} & \multicolumn{6}{c}{Instance Segmentation} \\
& & & & \multicolumn{1}{c}{AP$_{50}$} & \multicolumn{1}{c}{AP$_{75}$} & \multicolumn{1}{c}{AP} & \multicolumn{1}{c}{AP$_{\text{S}}$} & \multicolumn{1}{c}{AP$_{M}$} & \multicolumn{1}{c|}{AP$_{\text{L}}$} 
& \multicolumn{1}{c}{AP$_{50}$} & \multicolumn{1}{c}{AP$_{75}$} & \multicolumn{1}{c}{AP} & \multicolumn{1}{c}{AP$_{\text{S}}$} & \multicolumn{1}{c}{AP$_{M}$} & \multicolumn{1}{c}{AP$_{\text{L}}$} \\ [.1em]
\shline
FreeSOLO$^*$ & SOLOv2 & DenseCL & IN+COCO$^{\text{X+U}}$ & \underline{3.8}	&	\underline{1.6}	&	\underline{1.9}	&	\underline{0.8}	&	\underline{3.2}	&	\underline{6.8} & \underline{3.6}	&	\underline{1.7}	&	\underline{1.9}	&	\underline{0.5}	&	\underline{2.7}	&	\underline{7.3}\\
\hline
\Ours & Cascade & DINO & IN & 8.7	&	4.2	&	4.6	&	2.4	&	9.6	&	16.0 & 7.1	&	3.5	&	3.9	&	1.6	&	5.9	&	14.2 \\
$\Delta$ &&&& \plus{+4.9}	&	\plus{+2.6}	&	\plus{+2.7}	&	\plus{+1.6}	&	\plus{+6.4}	&	\plus{+9.2} & \plus{+3.5}	&	\plus{+1.8}	&	\plus{+2.0}	&	\plus{+0.9}	&	\plus{+3.2}	&	\plus{+6.9} \\
\end{tabular}
\end{center}
\vspace{-4mm}
\caption{Unsupervised multi-object detection and instance segmentation on \textbf{LVIS \texttt{val}}. $^*$: reproduced results with official codes and released checkpoint.}
\label{tab:lvis}
\end{table*}
}

\def\tabFineTuneMSCOCO#1{
\begin{figure*}[#1]
  \centering
  \begin{tabular}{c|c}
  Mask R-CNN & Cascade Mask R-CNN  \\
  \includegraphics[width=0.48\linewidth]{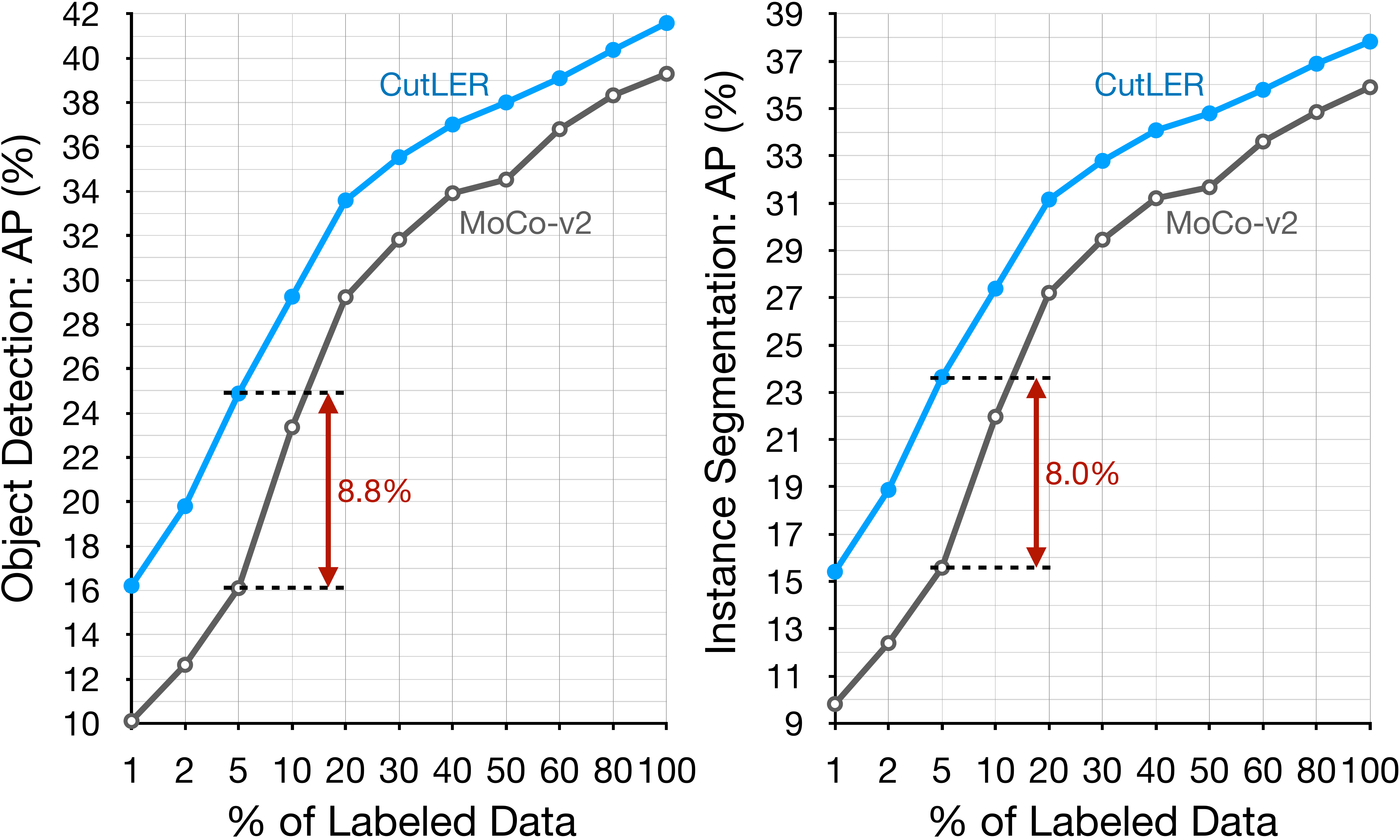}&
  \includegraphics[width=0.48\linewidth]{figures/SSL_Cascade.pdf}\\
  \end{tabular}
  \caption{Fine-tuning on MS-COCO with various annotation ratios. We report results using Mask R-CNN and Cascade Mask R-CNN with a backbone of ResNet-50 as the detector.
  }
  \label{fig:fine-tune}
\end{figure*}
}

\def\tabDatasets#1{
  \begin{table*}[#1]
  \tablestyle{5pt}{1.0}
  \small
  \begin{center}
  \begin{tabular}{lccccc}
  datasets & domain & testing data & \#images & instance segmentation label \\ [.1em]
  \shline
  COCO~\cite{lin2014microsoft} & natural images & \texttt{val2017} split & 5,000 & \cmark \\
  COCO20K~\cite{lin2014microsoft} & natural images & a subset of COCO & 20,000 & \cmark \\
  UVO~\cite{wang2021unidentified} & video frames & \texttt{val} split & 7,356 & \cmark \\
  LVIS~\cite{gupta2019lvis} & natural images & \texttt{val} split & 19,809 & \cmark \\
  KITTI~\cite{geiger2012we} & traffic images & \texttt{trainval} split & 7,521 & \xmark \\
  Pascal VOC~\cite{everingham2010pascal} & natural images & \texttt{trainval07} split & 9,963 & \xmark \\
  Clipart~\cite{inoue2018cross} & clip arts & \texttt{traintest} split & 1,000 & \xmark \\
  Watercolor~\cite{inoue2018cross} & paintings & \texttt{traintest} split & 2,000 & \xmark \\
  Comic~\cite{inoue2018cross} & sketches & \texttt{traintest} split & 2,000 & \xmark \\
  Objects365-V2~\cite{shao2019objects365} & natural images & \texttt{val} split & 80,000 & \xmark \\
  OpenImages-V6~\cite{kuznetsova2020open} & natural images & \texttt{val} split & 41,620 & \xmark \\
  \shline
  \end{tabular}
  \end{center}
  \vspace{-2pt}
  \caption{Summary of datasets used for zero-shot evaluation.}
  \label{tab:dataset-summary}
  \end{table*}
}

\def\tabResultsAll#1{
  \begin{table*}[#1]
  \tablestyle{1pt}{1.0}
  \small
  \begin{center}
  \begin{tabular}{l|ccccccccc|ccccccccc}
  \multirow{1}{*}{Datasets} & \multicolumn{1}{c}{AP$_{50}^{\text{box}}$} & \multicolumn{1}{c}{AP$_{75}^{\text{box}}$} & \multicolumn{1}{c}{AP$^{\text{box}}$} & \multicolumn{1}{c}{AP$_{S}^{\text{box}}$} & \multicolumn{1}{c}{AP$_{M}^{\text{box}}$} & \multicolumn{1}{c}{AP$_{L}^{\text{box}}$} & \multicolumn{1}{c}{AR$_{1}^{\text{box}}$} & \multicolumn{1}{c}{AR$_{10}^{\text{box}}$} & \multicolumn{1}{c|}{AR$_{100}^{\text{box}}$} & \multicolumn{1}{c}{AP$_{50}^{\text{mask}}$} & \multicolumn{1}{c}{AP$_{75}^{\text{mask}}$} & \multicolumn{1}{c}{AP$^{\text{mask}}$} & \multicolumn{1}{c}{AP$_{S}^{\text{mask}}$} & \multicolumn{1}{c}{AP$_{M}^{\text{mask}}$} & \multicolumn{1}{c}{AP$_{L}^{\text{mask}}$} & \multicolumn{1}{c}{AR$_{1}^{\text{mask}}$} & \multicolumn{1}{c}{AR$_{10}^{\text{mask}}$} & \multicolumn{1}{c}{AR$_{100}^{\text{mask}}$} \\ [.1em]
  \shline
  COCO & 21.9	&	11.8	&	12.3	&	3.7	&	12.7	&	29.6 & \phantom{1}6.8 & 19.6 & 32.8 & 18.9	&	\phantom{1}9.2	&	\phantom{1}9.7	&	2.4	&	\phantom{1}8.8	&	24.3 & 5.8 & 16.5 & 27.1 \\
  COCO20K & 22.4 & 11.9 & 12.5 & 4.1 & 12.7 & 29.5 & \phantom{1}6.8 & 19.7 & 33.1 & 19.6	&	\phantom{1}9.2	& 10.0 & 2.8 & \phantom{1}8.9 & 24.3 & 5.8 & 16.6 & 27.4 \\
  UVO & 31.7	&	14.1	&	16.1	&	3.7	&	11.3	&	25.3 & \phantom{1}6.8 & 24.5 & 42.5 & 31.6	&	14.1	&	16.1	&	3.7	&	11.3	&	25.3 & 4.6 & 18.0 & 32.2 \\
  LVIS & \phantom{1}8.4	&	\phantom{1}3.9	&	\phantom{1}4.5	&	2.7	&	\phantom{1}9.1	&	15.1 & \phantom{1}2.4 & \phantom{1}9.2 & 21.8 & \phantom{1}6.7	&	\phantom{1}3.2	&	\phantom{1}3.5	&	1.9	&	\phantom{1}6.1	&	12.5 & 2.1 & \phantom{1}7.9 & 18.7 \\
  KITTI & 18.4	&	\phantom{1}6.7	&	\phantom{1}8.5	&	0.5	&	\phantom{1}5.6	&	19.2 & \phantom{1}6.2 & 16.6 & 27.8 &-&-&-&-&-&-&-&-&- \\
  Pascal VOC & 36.9	&	19.2	&	20.2	&	1.3	&	\phantom{1}6.5	&	32.2 & 16.5 & 32.8 & 44.0 &-&-&-&-&-&-&-&-&-\\
  Clipart & 21.1	&	\phantom{1}6.0	&	\phantom{1}8.7	&	1.1	&	\phantom{1}5.8	&	11.6 & \phantom{1}6.6 & 27.0 & 40.7 &-&-&-&-&-&-&-&-&-\\
  Watercolor & 37.5	&	10.9	&	15.7	&	0.1	&	\phantom{1}1.1	&	20.0 & 19.4 & 37.8 & 44.2 &-&-&-&-&-&-&-&-&-\\
  Comic & 30.4	&	\phantom{1}7.7	&	12.2	&	0.0	&	\phantom{1}1.3	&	16.0 & \phantom{1}8.5 & 28.2 & 38.4 &-&-&-&-&-&-&-&-&-\\
  Objects365 & 21.6	&	10.3	&	11.4	&	3.0	&	10.4	&	20.4 & \phantom{1}3.0 & 15.4 & 34.2 &-&-&-&-&-&-&-&-&-\\
  OpenImages & 17.3	&	\phantom{1}9.5	&	\phantom{1}9.7	&	0.4	&	\phantom{1}2.3	&	14.9 & \phantom{1}6.5 & 17.6 & 29.6 &-&-&-&-&-&-&-&-&-\\
  \shline
  \end{tabular}
  \end{center}
  \caption{Detailed zero-shot evaluation results on all benchmarks used in this work.}
  \label{tab:results-all}
  \end{table*}
}

\def\tabExtraVis#1{
\begin{figure*}[#1]
  \centering
  \includegraphics[width=0.8\linewidth]{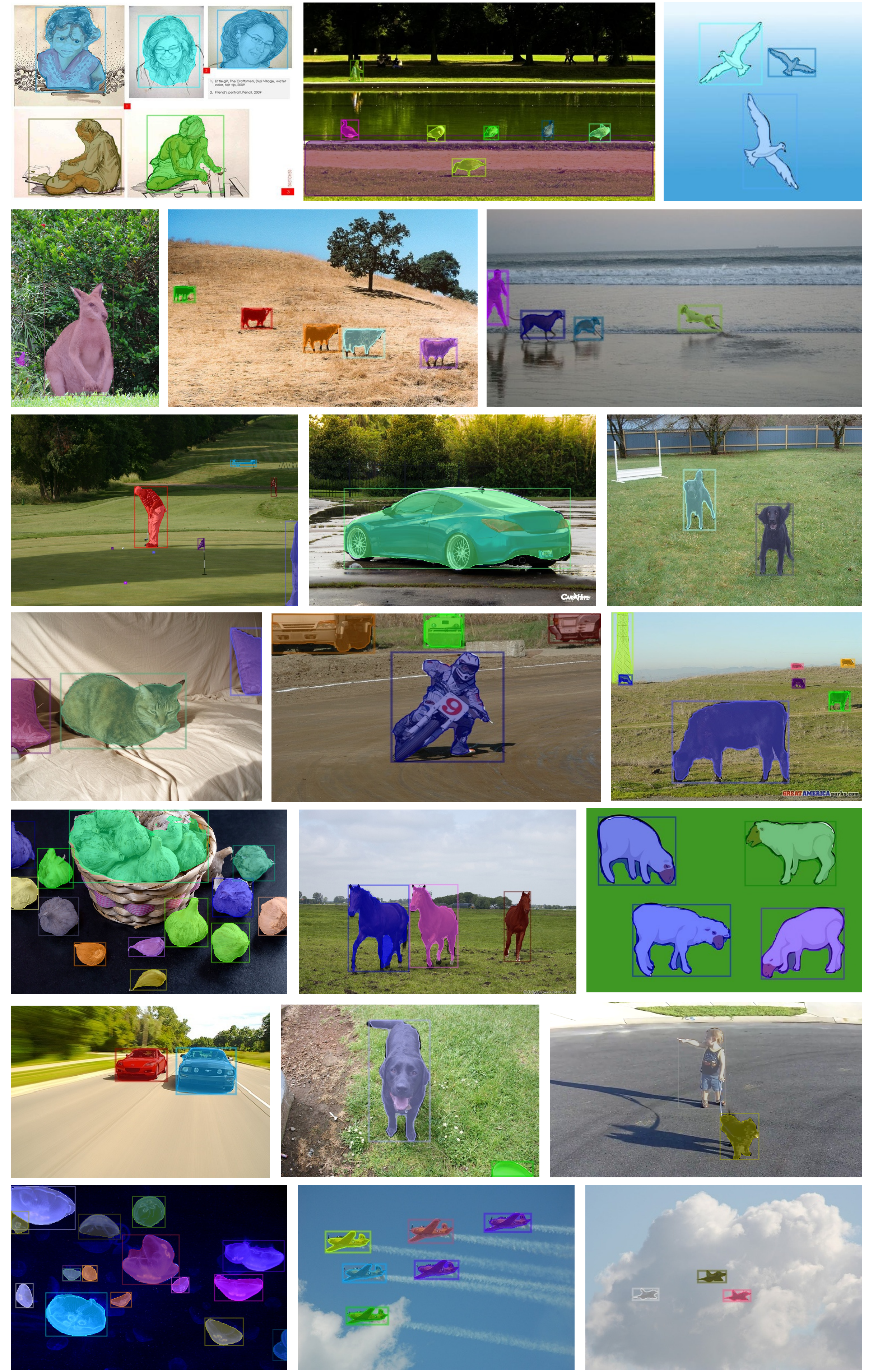}
  \caption{More visualizations of \Ours's predictions.}
  \label{fig:extra-visual}
\end{figure*}
}

\clearpage

\appendix
\section{Appendix}
\subsection{Training details}
\label{appendix:training-details}

\tabDatasets{t!}

While \Ours is agnostic to the underlying detector, we use popular Mask R-CNN~\cite{he2017mask} and Cascade Mask R-CNN~\cite{cai2018cascade} for all experiments, and use Cascade Mask R-CNN by default, unless otherwise noted.
We train the detector on \imnet with initial masks and bounding boxes for $160$K iterations with a batch size of 16.
When training the detectors with a ResNet-50 backbone~\cite{he2016deep}, we initialize the model with the weights of a self-supervised pretrained DINO~\cite{caron2021emerging} model.
We explored other pre-trained models, including MoCo-v2~\cite{chen2020improved}, SwAV~\cite{caron2020unsupervised}, and CLD~\cite{wang2021unsupervised}, and found that they give similar detection performance.
Therefore, we initialize model weights with DINO by default.

We also leverage the copy-paste augmentation \cite{ghiasi2021simple,dwibedi2017cut} during the model training process. Rather than using the vanilla copy-paste augmentation to improve the model's ability to segment small objects, we randomly downsample the mask with a scalar uniformly sampled between 0.3 and 1.0.
We then optimize the detector for $160K$ iterations using SGD with a learning rate of 0.005, which is decreased by 5 after $80K$ iterations and a batch size of 16.
We apply a weight decay of $5\!\times\!10^{-5}$ and a momentum of 0.9. 

For the multi-round of self-training, in each stage, we initialize the detection model using the weights from the previous stage. We optimize the detector using SGD with a learning rate of 0.01 for 80$K$ iterations. Since the self-training stage can provide a sufficient number of pseudo-masks for model training, we don't use the exploration loss during the self-training stage.

\subsection{Datasets used for zero-shot evaluation}
\label{appendix:zero-shot-dataset-details}

\noindent\textit{\textbf{COCO and COCO20K}}~\cite{lin2014microsoft} is a large-scale object detection and instance segmentation dataset, containing about 115$K$ and 5$K$ images in the training and validation split, respectively. Additionally, COCO has an unannotated split of 123$K$ images. We test our model in a class-agnostic manner on COCO \texttt{val2017} and COCO \texttt{20K}, without fine-tuning on any images in COCO. 
COCO 20$K$ is a subset of the COCO \texttt{trainval2014}~\cite{lin2014microsoft}, containing 19817 randomly sampled images, used as a benchmark in~\cite{vo2020toward,simeoni2021localizing, wang2022tokencut}.
We report class-agnostic COCO style averaged precision and averaged recall for object detection and segmentation tasks. 

\noindent\textit{\textbf{Pascal VOC}}~\cite{everingham2010pascal} is another popular benchmark for object dtetection. We evaluate our model on its \texttt{trainval07} split in COCO style evaluation matrics. 

\noindent\textit{\textbf{UVO}}~\cite{wang2021unidentified}. Unidentified Video Objects (UVO) is an exhaustively annotated dataset for video object detection and instance segmentation. 
We evaluate our model on UVO \texttt{val} by frame-by-frame inference and report results in COCO style evaluation matrics. 

\noindent\textit{\textbf{LVIS}}~\cite{gupta2019lvis} collected 2.2 million high-quality instance segmentation masks for over 1000 entry-level object categories, which naturally constitutes the long-tailed data distribution. We report class-agnostic object detection and instance segmentation results on LVIS \texttt{val} split, containing about 5$K$ images.

\noindent\textit{\textbf{CrossDomain}}~\cite{inoue2018cross} contains three subsets of watercolor, clipart, and comics, in which objects are depicted in watercolor, sketch and painting styles, respectively. We evaluate our model on all annotated images from these three datasets, \ie, \texttt{traintest}.

\noindent\textit{\textbf{Objects365}} \textit{V2}~\cite{shao2019objects365} presents a supervised object detection benchmark with a focus on diverse objects in the wild. We evaluate \Ours on the 80K images from its \texttt{val} split.

\noindent\textit{\textbf{OpenImages}} \textit{V6}~\cite{kuznetsova2020open} unifies image classification, object detection, and instance segmentation, visual relationship detection, \etc in one dataset. We evaluate \Ours on its ~42K images from the \texttt{val} split.

\noindent\textit{\textbf{KITTI}}~\cite{geiger2012we} presents a dataset captured from cameras mounted on mobile vehicles used for autonomous driving research. We evaluate \Ours on 7521 images from KITTI's \texttt{trainval} split. 

We provide the summary of these datasets used for zero-shot evaluation in~\cref{tab:dataset-summary}.

\tabResultsAll{t!}
\subsection{Additional results for zero-shot detection \& segmentation}
\label{appendix:zero-shot-additional}
In this section, we use official COCO API and provide more results with standard COCO metrics, including AP across various IoU thresholds - AP (averaged over IoU thresholds from 0.5 to 0.95 with a step size of 0.05), AP$_{50}$ (IoU@$0.5$) and AP$_{75}$ (IoU@$0.75$), and AP across scales - AP$_{\text{S}}$ (small objects), AP$_{\text{M}}$ (medium objects) and AP$_{\text{L}}$ (large objects). 
We provide detailed results on all these benchmarks listed in~\cref{tab:dataset-summary} and report these results in~\cref{tab:results-all}. 
We report the performance of object detection for all datasets.
In addition, for those datasets that provide annotations for instance segmentation, we also present the performance of the instance segmentation task. 
It is worth noting that on these datasets without segmentation labels, \Ours can still predict instance segmentation masks, but since we do not have ground truth masks to be compared, we cannot evaluate the results.

\subsection{\Ours vs. Selective Search}
\label{appendix:selective-search}

Selective Search~\cite{uijlings2013selective} is a popular unsupervised object discovery method, used in many early state-of-the-art detectors such as R-CNN \cite{girshick2014rich} and Fast R-CNN \cite{girshick2015fast}. 
However, generating possible object locations with sliding windows greatly reduces inference speed (please refer to \cite{uijlings2013selective} for more details on selective search).
We compare \Ours's performance to selective search in~\cref{fig:selective-search} and observe that \Ours provides a significant improvement in both precision and recall, which indicates that \Ours is a better performing unsupervised method for region proposal generation with real-time inference speed.

\tabSelectiveSearch{t!}

\tabFineTuneMSCOCO{t!}

\subsection{Training details for label-efficient and fully-supervised learning}

We train the detector on the COCO~\cite{lin2014microsoft} dataset using the bounding box, and instance mask labels.
To evaluate label efficiency, we subsample the training set to create subsets with varying proportions of labeled image
We train the detector, initialized with \Ours, on each of these subsets.

As a baseline, we follow the settings from MoCo-v2~\cite{chen2020improved} and train the same detection architecture initialized with a MoCo-v2 ResNet50 model, given its strong performance on object detection tasks.
MoCo-v2 and our models use the same training pipeline and hyper-parameters and are trained for the $1\times$ schedule using Detectron2~\cite{wu2019detectron2}, except for extremely low-shot settings with 1\% or 2\% labels. Following previous works~\cite{wang2022freesolo}, when training with 1\% or 2\% labels, we train both MoCo-v2 and our model for 3,600 iterations with a batch size of 16. 

Our detector weights are initialized with ImageNet-1K pre-trained \Ours, except for the weights of the final bounding box prediction layer and the last layer of the mask prediction head, which are randomly initialized with values taken from a normal distribution. 
For experiments on COCO with labeling ratios below 50\%, during model training, we use a batch size of 16, and learning rates of 0.04 and 0.08 for model weights loaded from the pre-trained \Ours and randomly initialized, respectively.
For experiments on COCO with labeling ratios between 50\% and 100\%, the learning rates of all layers decay by a factor of 2.

For a fair comparison, baselines and \Ours use the same hyper-parameters and settings.

\subsection{More visualizations}
We provide more qualitative visualizations of \Ours's zero-shot predictions in~\cref{fig:extra-visual}.

\tabExtraVis{t!}

\end{document}